\documentclass{article} 
\let\hropeaddcontentsline\addcontentsline
\usepackage{iclr2027_conference,times}
\let\addcontentsline\hropeaddcontentsline

\usepackage{amsmath,amsfonts,bm}

\def\eqref#1{equation~\ref{#1}}

\def\1{\bm{1}}

\DeclareMathAlphabet{\mathsfit}{\encodingdefault}{\sfdefault}{m}{sl}
\SetMathAlphabet{\mathsfit}{bold}{\encodingdefault}{\sfdefault}{bx}{n}

\usepackage{bm}

\usepackage[normalem]{ulem}
\usepackage{xcolor}
\usepackage{multirow}
\usepackage{wrapfig,lipsum,booktabs}
\usepackage{listings}
\usepackage{algpseudocode}
\usepackage{amssymb}
\usepackage{tocloft}
\tocloftpagestyle{fancy}
\usepackage{setspace}
\usepackage{lineno}
\definecolor{paramRT}{RGB}{31, 155, 26}
\definecolor{paramKappa}{RGB}{191,78,110}
\definecolor{paramN}{RGB}{52,88,207}

\DeclareRobustCommand{\paramcell}[2]{%
  \begingroup
  \setlength{\fboxsep}{1.2pt}%
  \mathord{\colorbox{#1!12!white}{%
    \raisebox{0pt}%
      [\dimexpr\height+0.6pt\relax]%
      [\dimexpr\depth+1.5pt\relax]{%
      \makebox[1.2em][c]{%
        $\textstyle\color{#1}
         \vphantom{\mathbf{n}^{\top}}#2$}}}}%
  \endgroup
}
\usepackage{hyperref}
\usepackage{url}
\usepackage{enumitem} 
\usepackage{graphicx} 
\usepackage{multirow} 
\usepackage{colortbl,xcolor} 
\usepackage{algorithm}       
\usepackage{algpseudocode}   
\usepackage{tabularx}

\renewcommand{\eqref}[1]{\textup{\ref{#1}}}
\usepackage[table]{xcolor}   
\definecolor{besthl}{RGB}{228,240,215}
\definecolor{sndhl}{RGB}{206,221,234}

\usepackage{xspace}
\newcommand{\ourpe}{HRoPE\xspace}

\title{HRoPE: Homography Relative Positional Encoding}

\author{%
\parbox[t]{\dimexpr\textwidth-2\tabcolsep\relax}{%
\raggedright\normalfont
\textbf{Huachen Gao}\textsuperscript{1,2,}\thanks{These authors contributed equally.}\quad
\textbf{Yang Yang}\textsuperscript{3,*}\quad
\textbf{Xiaoqian Liang}\textsuperscript{2}\quad
\textbf{Xingyu Chen}\textsuperscript{1,2}\\
\textbf{Youjia Zhang}\textsuperscript{3}\quad
\textbf{Qi Zhang}\textsuperscript{4}\quad
\textbf{Baorui Ma}\textsuperscript{5}\quad
\textbf{Anpei Chen}\textsuperscript{2,}\thanks{Corresponding authors.}\quad
\textbf{Wei Yang}\textsuperscript{3,\textdagger}\\[3pt]
{\normalfont
\textsuperscript{1}Zhejiang University\quad
\textsuperscript{2}Westlake University\\
\textsuperscript{3}Huazhong University of Science and Technology\\
\textsuperscript{4}VIVO\quad
\textsuperscript{5}Tsinghua University}
}
}

\newcommand{\best}[1]{\cellcolor{besthl}#1}
\newcommand{\snd}[1]{\cellcolor{sndhl}#1}

\iclrfinalcopy 
\begin{document}

\maketitle
\lhead{Preprint}

\begin{abstract}
Geometric Relative Positional Encoding has become the primary mechanism for injecting spatial inductive biases into multi-view transformers. Existing designs condition attention on camera-level frustums or ray-marched depth, associating cross-view tokens through the sensor geometry. Such sensor-centric formulations, however, describe only where a token is observed from, leaving the rich surface geometry of the scene region depicted by the token.
In this paper, we advocate a shift toward surface-aware positional encoding that explicitly grounds patch tokens in the geometry of physical surfaces. We introduce \textbf{\ourpe}, which represents the local geometry of each patch as a slanted 3D plane, following the slanted support-window assumption of classical multi-view stereo.
We inject token-level, plane-induced homographies into attention as relative transformations. This equips the transformer with a geometric prior for rectifying perspective distortions and aligning patch features across viewpoints. 
Our encoding requires no ground-truth geometry: the plane parameters can be predicted end-to-end under a view-synthesis objective alone, or derived from foundation models. Extensive experiments on RealEstate10K, CO3Dv2, and DL3DV show that our best variant of \ourpe{} consistently outperforms the evaluated geometric positional encodings for novel view synthesis. Extensive ablations and analysis show the effectiveness of the proposed encoding. Please refer to our project page at: \href{https://gaohchen.github.io/HRoPE/}{https://gaohchen.github.io/HRoPE}.

\end{abstract}

\section{Introduction}

The Transformer has become the de facto architecture for recent feed-forward 3D vision~\citep{wang2024dust3r, vggt, lvsm}. By tokenizing images from multiple viewpoints and exchanging information via attention, these models exhibit strong capabilities in jointly resolving appearance, visibility, and scene geometry. Attention, however, is permutation-equivariant: it operates on unordered sets and is inherently agnostic to the camera poses, metric scales, and epipolar constraints that couple multi-view observations. Consequently, geometric priors must be explicitly injected, either by conditioning token representations directly (e.g., via camera tokens~\citep{vggt} or ray maps~\citep{lvsm, gslrm}) or by imposing epipolar constraints on attention~\citep{he2020epipolar, pixelsplat, ye2024consistent}. Positional encoding (PE) introduces geometric structure directly into attention by modulating pairwise token comparisons.

\begin{figure}[t]
\centering
\includegraphics[width=\linewidth]{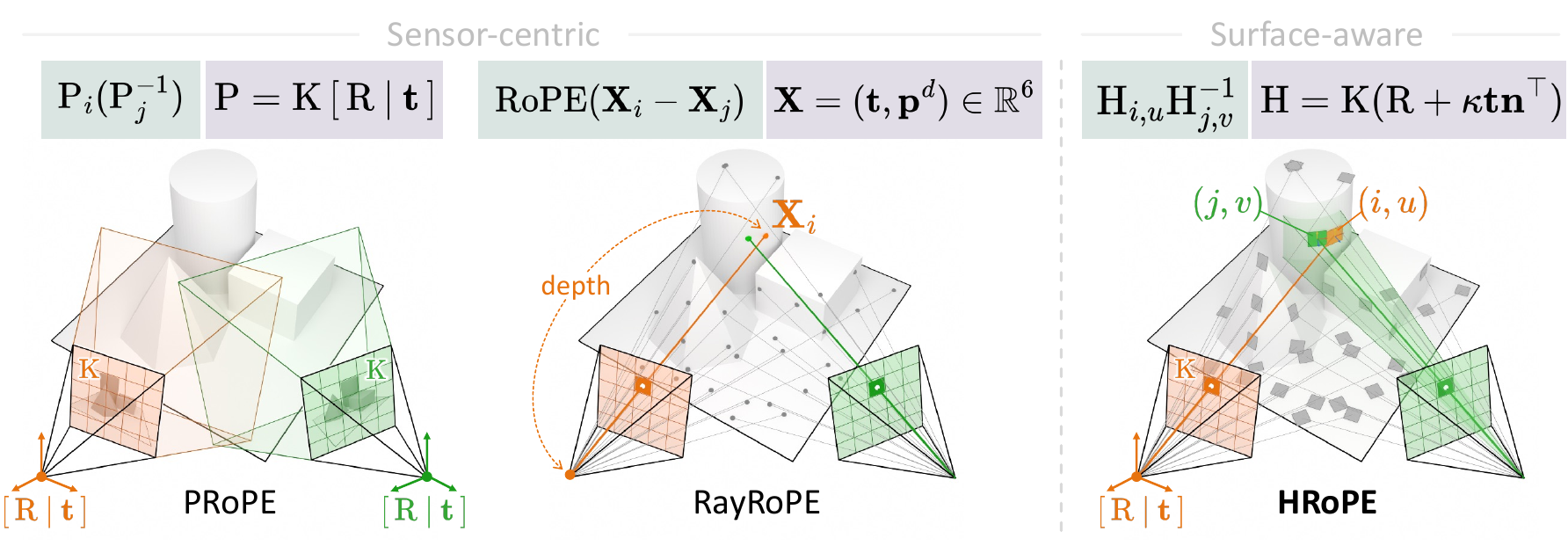}
\caption{\textbf{From sensor-centric to surface-aware geometric positional encoding.}
\textbf{Left:} PRoPE~\citep{prope} assigns every token in a view the same camera
projection $\mathrm{P}=\mathrm{K}[\mathrm{R}\mid\mathbf{t}]$.
\textbf{Middle:} RayRoPE~\citep{rayrope} predicts a depth $d$ for each token and
places it at a 3D point $\mathbf{X}$ along the camera ray, but does not model local
surface orientation. \textbf{Right:} \ourpe{} associates each patch with a slanted
tangent plane and relates token pairs through the plane-induced homography
$\mathrm{H}_{i,u}\mathrm{H}_{j,v}^{-1}$ (Eq.~\ref{eq:homography}), thereby encoding
both depth and local surface orientation in the attention comparison.}
\label{fig:teaser}
\end{figure}

\vspace{2pt}
While existing PE designs exhibit increasing geometric fidelity by using priors from images, cameras, and rays, they invariably anchor positional information to the sensor.
RoPE extensions~\citep{su2024roformer} encode 2D intra-view offsets, leaving inter-view transformations implicit~\citep{weinzaepfel2023croco, wang2024dust3r}. GTA~\citep{gta} and CaPE~\citep{cape} inject relative camera extrinsics to achieve invariance to global rigid motions, while PRoPE~\citep{prope} generalizes this to the full projection matrix. Ray-based formulations like RayRoPE~\citep{rayrope} and URoPE~\citep{urope} integrate depth by localizing patches along viewing rays and re-projecting into the query frame. Across this progression, geometry is parameterized via sensor-centric abstractions: an image coordinate, a camera frustum, or a point of ray termination, leaving the rich local surface structure of the underlying scene unmodeled (see Fig.~\ref{fig:teaser}, left and middle).

\vspace{2pt}
This limitation is fundamental because multi-view matching does not compare isolated points; it compares local neighborhoods whose appearances are related through surface-induced deformations. A patch in one view and its corresponding patch in another are connected by a warp determined by the underlying surface geometry, and reliable correspondence requires compensating for this warp before comparing visual features. Given the pixel location and camera geometry, depth locates the patch center, while surface orientation additionally determines how its surrounding neighborhood deforms across views. By providing only view or positional information, existing PEs implicitly impose a fronto-parallel assumption on the attention computation mechanism. Although large neural models can learn to compensate for this bias given sufficient training data, doing so compromises efficiency and generalization. We argue that surface orientation is the missing degree of freedom: current PEs capture the zeroth-order position of a patch but neglect the first-order local orientation that dictates its cross-view appearance transformation.

\vspace{2pt}
To this end, we present \textbf{\ourpe{}}, a Homography Relative Positional Encoding that embeds orientation-aware rectification directly into the attention mechanism (Fig.~\ref{fig:teaser}, right). Inspired by the slanted support windows of classical multi-view stereo~\citep{patchmatch, patchmatchnet}, \ourpe{} anchors a local tangent plane to each patch token. We use the plane normal and distance, together with the camera poses, to define the homography governing cross-view neighborhood deformations. Concretely, we apply a per-patch $3\times3$ transformation block-diagonally to the queries, keys, and values, and incorporate a closed-form inverse into the key projection so that the resulting attention logits depend exclusively on the relative homography between tokens. As we show in Sec.~\ref{sec:homography} and Appendix~\ref{app:derivation}, this mechanism acts as an in-attention analogue of computing a matching score only after both neighborhoods have been rectified onto a shared tangent plane. Consequently, whereas prior geometric PEs merely encode where a token is located, \ourpe{} explicitly models how its underlying surface transforms across views. This formulation offers two compelling advantages. First, \ourpe{} strictly generalizes point- and ray-based PEs: when a predicted normal aligns with the viewing direction, the induced transformation naturally reduces to the fronto-parallel case. Second, the per-patch transformation parameters can be regressed
from patch features by a lightweight head trained jointly with
the view-synthesis model, using the view-synthesis objective alone.
The head can optionally use pretrained features from
DINOv2~\citep{dinov2} or VGGT~\citep{vggt}, without changing the
encoding formulation or training objective.



\vspace{2pt}
We validate \ourpe{} on feed-forward novel view synthesis (NVS) across the RealEstate10K~\citep{real10k}, CO3Dv2~\citep{co3d}, and DL3DV~\citep{dl3dv} datasets, where it consistently outperforms view-, projection-, and ray-based encodings. Notably, \ourpe{} surpasses PRoPE on RealEstate10K without ground-truth camera poses, demonstrating that per-patch homography is a stronger prior than view-level projection. Furthermore, our ablations show that this geometry must condition token interactions multiplicatively, injecting the same information as additive features sharply degrades performance. Finally, \ourpe{} improves generalization to unseen zooms and wider baselines, supporting its role as a geometric inductive bias. Beyond NVS, \ourpe{} transfers unchanged to a multi-view diffusion model and to the visual tokens of a spatial reasoning MLLM, improving performance on both
tasks (Appendix~\ref{app:transfer}).

\vspace{2pt}
We summarize our main contributions as follows:
\begin{enumerate}[leftmargin=1.4em,itemsep=0.15em,topsep=0.2em]
\item We formulate \textit{the principle of surface-aware positional encoding}, positioning local 3D surface geometry as the natural anchor for comparing patch neighborhoods across views.



\item We propose \textbf{\ourpe{}}, a per-patch relative encoding that uses plane-induced transforms with closed-form inverses to condition attention logits on relative cross-view homographies.


\item We demonstrate consistent improvements over the evaluated geometric positional encodings on multiple NVS benchmarks and under out-of-distribution view settings. Controlled ablations analyze the roles of transformation structure, local geometry, and feature source, supporting the benefit of the homography inductive bias.


\end{enumerate}

\section{Related Work}
\paragraph{Geometric Positional Encoding.}
Geometric PEs encode camera or scene geometry in tokens
and attention. Absolute variants encode Pl\"ucker rays
\citep{sitzmann2021lfn,sajjadi2022srt,lvsm} or 3D positions~\citep{liu2022petr},
but depend on the chosen world frame. Relative variants extend rotary encodings
\citep{su2024roformer,shaw2018relative} to camera geometry: GTA~\citep{gta} and
CaPE~\citep{cape} use relative extrinsics, while PRoPE~\citep{prope} also includes
intrinsics through projection matrices. Recent methods move from view- to token-level
geometry: RayRoPE~\citep{rayrope} predicts per-token depth, RoRE~\citep{rore} rotates
features along rays, and URoPE~\citep{urope} projects per-head depth anchors into the
query frame, following epipolar attention~\citep{he2020epipolar,suhail2022light}. These encodings describe camera relationships and point- or ray-level geometry. \ourpe{} adds an explicit local-plane parameterization whose orientation, together with distance and camera geometry, determines a planar neighborhood’s cross-view deformation. This distinguishes surface-conditioned transformations from point localization without requiring the latter to assume fronto-parallel surfaces.


\paragraph{Multi-view Transformers.}
Multi-view transformers use camera geometry to associate image tokens across
views. Their applications include stereo depth
estimation~\citep{he2020epipolar,yao2018mvsnet,gu2020cascade}, camera and structure
recovery~\citep{wang2024dust3r,vggt}, cross-view completion and
matching~\citep{weinzaepfel2023croco}, spatial reasoning~\citep{vsibench,spatialmllm},
and camera-controlled video and world modeling~\citep{rerope,ucpe,ma2025you},
as well as novel view synthesis (NVS).
For feed-forward NVS, methods use feature-conditioned
radiance fields~\citep{pixelnerf,ibrnet,mvsnerf,trevithick2021grf,hong2024lrm},
predict 3D Gaussians~\citep{pixelsplat,mvsplat,gslrm}, or directly regress
pixels~\citep{lvsm,rayzer}.
We evaluate \ourpe{} primarily on novel view synthesis using the
LVSM backbone with plane-induced positional encoding.


\paragraph{Slanted Support Windows.}
Traditional local stereo methods assume constant disparity within
a support window, introducing a fronto-parallel bias on slanted
surfaces~\citep{patchmatch}.
Building on randomized correspondence search~\citep{barnes2009patchmatch}, \mbox{PatchMatch Stereo~\citep{patchmatch}} addresses this limitation by estimating
a local disparity plane at each pixel and evaluating matching costs
over slanted support windows.
More generally, local planar geometry induces a homography that relates
corresponding image neighborhoods across views~\citep{hartley2003multiview}.
Learned multi-view stereo methods also use differentiable geometric warping
to compare features under depth hypotheses~\citep{yao2018mvsnet,gu2020cascade,patchmatchnet}.
\ourpe{} draws on the principle of using local surface geometry to guide
cross-view matching. It incorporates per-patch, plane-induced transformations
into multi-view attention, bringing the geometric intuition of
slanted-window matching into positional encoding.


\section{Method}


\begin{figure}[t]
\centering
\includegraphics[width=\linewidth]{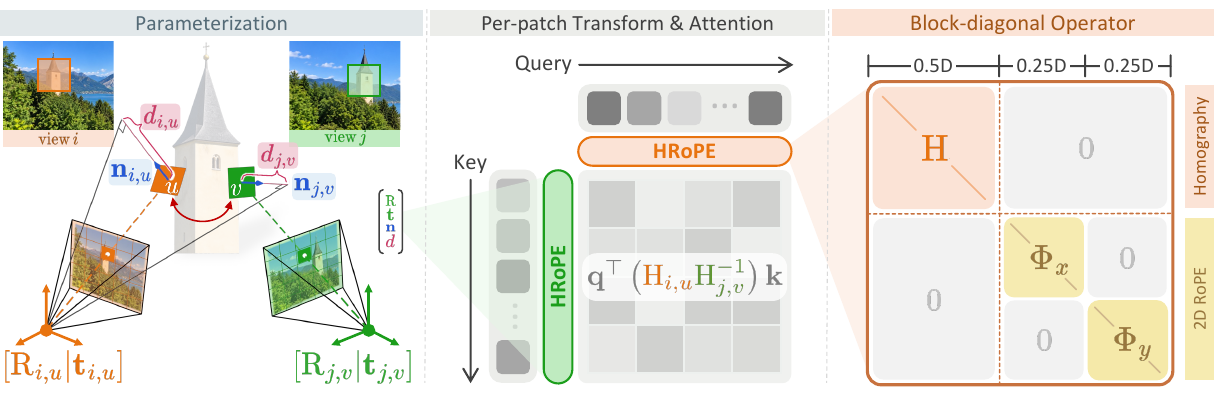}
\caption{\textbf{Overview of \ourpe{}.}
\textbf{Left:} per-patch parameters
$(\mathrm{R},\mathbf{t},\mathbf{n},\kappa)$ are predicted from
patch-level image features (Sec.~\ref{sec:source}),
where $\mathbf{n}$ and $\kappa=1/d$ represent the local plane's
unit normal and inverse plane distance, respectively.
Together with intrinsics $\mathrm{K}$, these parameters define
\ensuremath{
  \mathrm{H}=\mathrm{K}\bigl(
  \paramcell{paramRT}{\mathrm{R}}
  +
  \paramcell{paramKappa}{\kappa}\mkern1mu
  \paramcell{paramRT}{\mathbf{t}}\mkern1mu
  \paramcell{paramN}{\mathbf{n}^{\top}}
  \bigr)
}
(Eq.~\eqref{eq:A}).
 \textbf{Middle:} token-specific transforms modulate the cross-view attention, yielding the relative factor
$\mathrm{H}_{i,u}\mathrm{H}_{j,v}^{-1}$ in each attention logit
(Eq.~\eqref{eq:logit}). \textbf{Right:} the transform is repeated block-diagonally
over the geometric half of each head, while the remaining dimensions retain the
$2$D rotary blocks $\Phi_x$ and $\Phi_y$ (Eq.~\eqref{eq:app-bigmat}).}
\label{fig:pipeline}
\end{figure}


\ourpe{} is a geometric positional encoding that incorporates a per-token surface plane transform.  In Sec.\ref{sec:prelim}, we first review how prior work incorporates camera geometry into multi-view attention. Then, in Sec.~\ref{sec:homography}, we introduce the plane-induced transformation and show how it modulates multi-view attention. With consistent camera geometry, the relative transformation exactly maps points on a shared physical plane across views. For independently predicted per-patch parameters, the same algebraic form modulates attention without requiring an exact physical correspondence for every token pair. Finally, Sec.~\ref{sec:source} details how these per-patch parameters are predicted from image features and presents a calibrated variant that leverages known camera parameters. The overall pipeline of \ourpe{} is illustrated in Fig.~\ref{fig:pipeline}.


\subsection{Preliminaries: Geometric Positional Encodings}
\label{sec:prelim}

Self-attention alone does not encode token positions or camera geometries. RoPE
\citep{su2024roformer} applies position-dependent rotations to queries and keys so
that their dot product depends on relative positions. Multi-view relative
PEs extend this principle to cameras: GTA~\citep{gta} and CaPE~\citep{cape}
encode relative extrinsics, while PRoPE~\citep{prope} adds intrinsics.
\paragraph{Projective Relative Encodings.}
Let $\mathrm{P}_i$ be PRoPE's $4\times4$ projection transform for view $i$,
built from its intrinsics and extrinsics. For a query $\mathbf{q}$ in view
$i$ and a key $\mathbf{k}$ in view $j$, the attention score is
$(\mathrm{P}_i^\top \mathbf{q})^\top
(\mathrm{P}_j^{-1}\mathbf{k})
= \mathbf{q}^\top \big(\mathrm{P}_i \mathrm{P}_j^{-1}\big)\,\mathbf{k}$.
Thus, the score depends on the projection transforms only through
$\mathrm{P}_i\mathrm{P}_j^{-1}$, which is invariant to a shared change of
world coordinates. Yet all tokens in a view share the same transform,
so this view-level modulation cannot encode local surface variation.

\paragraph{From Camera to Surface Geometry.}
RayRoPE~\citep{rayrope} and URoPE~\citep{urope} describe token-level geometry using
sensor- or point-centered coordinates along viewing rays. To model local surface
deformation, we associate each patch with a tangent plane that captures its position
and orientation. Composing the camera projection with a surface-induced transform then allows the positional encoding to account for how the patch's surrounding neighborhood deforms across views.

\subsection{Homography Relative Positional Encoding}
\label{sec:homography}

Suppose the local surface observed by an image patch lies on a plane $\mathbf{n}^\top\mathbf{X}=d$ in the world coordinate frame, with unit normal $\mathbf{n}$ and signed distance $d$. The projection of points $\mathbf{X}$ on this plane into view $i$ can be expressed as a linear mapping $\tilde{\mathbf{x}}_i\sim\mathrm{H}_i\mathbf{X}$, where
\begin{equation}
\label{eq:Hi}
\mathrm{H}_i
=
\mathrm{K}_i\left(\mathrm{R}_i+\frac{\mathbf{t}_i\mathbf{n}^\top}{d}\right).
\end{equation}
Applying this similarly for view $j$ and eliminating $\mathbf{X}$ yields the plane-induced homography~\citep{malis2007homography} between the two views:
\begin{equation}
\label{eq:homography}
\tilde{\mathbf{x}}_i\sim\mathrm{H}_i\mathrm{H}_j^{-1}\tilde{\mathbf{x}}_j.
\end{equation}
For visible points on the assumed plane, Eq.~\eqref{eq:homography}
determines a correspondence between the two views, whereas epipolar
geometry alone restricts the match to a line. When a local neighborhood
is approximately planar, the same homography describes its cross-view
deformation. This use of a local plane is related to the slanted support
windows in PatchMatch Stereo~\citep{patchmatch}.

\paragraph{From Homography to Positional Encoding.}
Eq.~\eqref{eq:homography} suggests a factorized form of a relative positional encoding: a per-view transformation to a common reference frame, composed pairwise such that only the relative transformation survives. PRoPE realizes this structure using the camera projection matrix $\mathrm{P}_i$, where the relative factor $\mathrm{P}_i\mathrm{P}_j^{-1}$ is shared globally by every token pair between views $i$ and $j$. Replacing $\mathrm{P}_i$ with the plane-induced $\mathrm{H}_i$ from Eq.~\eqref{eq:Hi} allows the relative factor to depend on both the local surface geometry and the camera poses. Furthermore, assigning a distinct plane to each patch makes this transformation specific to each individual token pair. 

We adopt this exact formulation: each patch $(i,u)$ is assigned a transformation $\mathrm{H}_{i,u}$ in the form of Eq.~\eqref{eq:Hi}, parameterized by its local plane, and a pair of patches is related by the composition $\mathrm{H}_{i,u}\mathrm{H}_{j,v}^{-1}$. When both patches correspond to the same physical plane, this product exactly recovers the homography in Eq.~\eqref{eq:homography}. In general, it acts as a projective transformation determined by the surface parameters of both patches. We use this projective factor to modulate query-key comparisons according to patch geometry. For patches on different planes, it is not guaranteed to represent a physical image warp.

\paragraph{Per-patch Transform and Attention.}
For patch token $u$ in view $i$, we define:
\begin{equation}
\label{eq:H}
\mathrm{H}_{i,u}
=
\mathrm{K}_i\left(
\mathrm{R}_{i,u}
+\kappa_{i,u}\,\mathbf{t}_{i,u}\mathbf{n}_{i,u}^{\top}
\right),
\end{equation}
where $(R_{i,u}, t_{i,u})$ parameterizes a patch-specific rotation and translation, $\mathbf{n}_{i,u}$ and $\kappa_{i,u}$ act as a unit plane normal and inverse plane distance. Although camera extrinsics are shared within a view, we relax this constraint when using the homography parameterization to modulate learned features. Predicting $(R_{i,u}, t_{i,u})$ per patch allows both the rotational component and the rank-one correction to adapt to local features, while retaining the structured form of a plane-induced transformation. These quantities therefore serve as latent transformation parameters optimized for view synthesis.

Following PRoPE, we multiply the query of patch $(i,u)$ by $\mathrm{H}_{i,u}^{\top}$ and the key of patch $(j,v)$ by $\mathrm{H}_{j,v}^{-1}$. These operations are applied block-diagonally over $3$-dimensional chunks of the attention head dimension. Consequently, the attention logit (omitting standard scaling) is given by:
\begin{equation}
  \vspace{2pt}
\label{eq:logit}
\big(\mathrm{H}_{i,u}^{\top}\mathbf{q}\big)^{\!\top}
\big(\mathrm{H}_{j,v}^{-1}\mathbf{k}\big)
=
\mathbf{q}^\top
\big(\mathrm{H}_{i,u}\mathrm{H}_{j,v}^{-1}\big)\mathbf{k}.
\vspace{2pt}
\end{equation}
The logits are normalized using the standard softmax.
The value and output transformations are detailed in
Appendix~\ref{app:integration}. Unlike PRoPE, where all token pairs between views $i$ and $j$
share the same relative factor, \ourpe{} uses token-specific transforms, allowing the relative factor to vary across token pairs. For two patches $u$ and $v$ in the same view, the geometric part of their attention score depends on $\mathrm{H}_{i,u}\mathrm{H}_{i,v}^{-1}$. This factor is the identity when their predicted transforms are equal; otherwise, it reflects the difference between those transforms. We apply standard 2D RoPE to the remaining feature channels to encode the patches' relative image positions.

\paragraph{Parameterization and Closed-form Inverse.}
Evaluating the key-side transformation in Eq.~\eqref{eq:logit} requires computing the inverse of every predicted $\mathrm{H}_{j,v}$. Dropping the patch and view indices for brevity, we can decompose the transformation as:
\begin{equation}
  \vspace{2pt}
\label{eq:A}
\mathrm{A}=\mathrm{R}+\kappa\,\mathbf{t}\mathbf{n}^{\top},
\qquad
\mathrm{H}=\mathrm{K}\mathrm{A},
\qquad
\mathrm{H}^{-1}=\mathrm{A}^{-1}\mathrm{K}^{-1}.
\end{equation}
The intrinsic-free $3\times3$ matrix $\mathrm{A}$ is parameterized by $13$ predicted scalars per patch: a translation vector $\mathbf{t}\in\mathbb{R}^3$, a rotation matrix $\mathrm{R} \in \mathrm{SO}(3)$ derived from a $6$D continuous representation~\citep{zhou2019continuity} via Gram--Schmidt orthogonalization, a unit normal vector $\mathbf{n}$ normalized to unit length, and the inverse plane distance $\kappa$. 

Because $\mathrm{A}$ is constructed as a rank-one update to a rotation matrix, its exact inverse can be computed analytically using the Sherman--Morrison identity~\citep{sherman1950adjustment}:
\begin{equation}
  \vspace{2pt}
\label{eq:sm}
\mathrm{A}^{-1}
=
\mathrm{R}^\top
-\frac{\kappa\,(\mathrm{R}^\top\mathbf{t})(\mathbf{n}^\top\mathrm{R}^\top)}
{1+\kappa\,\mathbf{n}^\top\mathrm{R}^\top\mathbf{t}}.
\end{equation}
This identity allows both $\mathrm{A}$ and $\mathrm{A}^{-1}$ to be computed efficiently in a single forward pass from the same parameters, completely bypassing the need for explicit matrix inversion routines. The denominator in Eq.~\eqref{eq:sm} is equal to $\det\mathrm{A}$ and vanishes exactly when the camera center lies on the predicted plane. To ensure the transformation remains numerically well-conditioned, we clamp $\mathbf{t}$, $\kappa$, and this denominator (see Appendix~\ref{app:clamp} for details).

\subsection{Instantiation}
\label{sec:source}

\paragraph{Predicted Instance.}
The patch-specific transformation described in Sec.~\ref{sec:homography} is constructed from $(\mathrm{R}, \mathbf{t}, \mathbf{n}, \kappa)$. These parameters are predicted by a lightweight regression head that maps patch features to the components of $\mathrm{A}$ ($13$ scalar values in the general case). The input features can be extracted from a predictor trained from scratch (acting as a learnable positional encoding), a pretrained 2D backbone (e.g., DINO), or a 3D-aware backbone (e.g., VGGT). Our primary implementation regresses these parameters from VGGT~\citep{vggt} features; VGGT’s pretraining on geometric prediction tasks provides a strong inductive bias for local surface estimation. In Sec.~\ref{sec:exp}, we compare this choice against a 2D self-supervised backbone and a randomly initialized predictor, keeping the encoding formulation otherwise unchanged. Note that these geometric parameters are predicted exclusively for observed input views.

\paragraph{Calibrated Instance.}
When camera calibration is available, we retain PRoPE's projection
$\mathrm{P}_i$ and predict only a local plane per patch, represented
by a frame in world coordinates:
\begin{equation}
\vspace{2pt}
\label{eq:PH}
\mathrm{F}_{i,u}
=
\begin{bmatrix}
\mathrm{R}_{\mathbf{n}_{i,u}} & z_{i,u}\mathbf{n}_{i,u} \\
\mathbf{0}^\top & 1
\end{bmatrix},
\qquad
\mathrm{H}^{\mathrm{cal}}_{i,u}=\mathrm{P}_i\mathrm{F}_{i,u},
\vspace{2pt}
\end{equation}
where $\mathrm{R}_{\mathbf{n}}$ is the minimal rotation aligning
$\mathbf{e}_3$ with $\mathbf{n}$, and $z_{i,u}$ is the signed
orthogonal plane offset.
Unlike the predicted $3\times3$ homography-structured transform,
this variant uses a $4\times4$ composition of camera projection
and local frame.
Applying the same relative-attention rule (Eq.~\eqref{eq:logit})
to $4$-dimensional feature blocks yields the relative factor
$\mathrm{P}_i(\mathrm{F}_{i,u}\mathrm{F}_{j,v}^{-1})\mathrm{P}_j^{-1}$,
combining relative local-frame geometry with calibrated camera
projections.
When all patches share a common frame, this reduces to
$\mathrm{P}_i\mathrm{P}_j^{-1}$, which is the same as PRoPE.

\section{Experiments}
\label{sec:exp}
We first evaluate \ourpe{} on novel view synthesis, comparing it
with existing geometric positional encodings within the
LVSM~\citep{lvsm} framework (Sec.~\ref{sec:nvs}).
We then ablate the transformation parameterization and parameter
source (Sec.~\ref{sec:ablation}), and analyze the role and quality of the predicted local
geometry (Sec.~\ref{sec:bridge}).
Additional experiments evaluate generalization under shifts in
camera intrinsics and view sampling (Appendix~\ref{app:ood}),
as well as transfer to a multi-view diffusion model and to the
visual tokens of a spatial reasoning MLLM
(Appendix~\ref{app:transfer}).

\subsection{Novel View Synthesis}
\label{sec:nvs}
\paragraph{Experimental Setup.}
We evaluate \ourpe{} on novel view synthesis by applying it to the decoder-only LVSM~\citep{lvsm} architecture. Each model takes two reference views as input and synthesizes a target view. Following PRoPE~\citep{prope}, we train downsized LVSM backbone variants ($\sim$25M parameters) with different positional encoding methods. We train and evaluate the models separately on CO3Dv2~\citep{co3d}, RealEstate10K~\citep{real10k}, and Objaverse~\citep{deitke2023objaverse}, and report PSNR, SSIM~\citep{wang2004ssim}, and LPIPS~\citep{zhang2018lpips}.
To evaluate scaling with the number of reference views, we additionally follow the RayZer~\citep{rayzer} protocol and train a larger LVSM backbone variant ($\sim$53M parameters) on DL3DV~\citep{dl3dv}. We evaluate the model with $2$, $4$, and $8$ reference views on the held-out DL3DV benchmark.

\vspace{3pt}
\begin{table*}[t]
\centering
\caption{Novel view synthesis on CO3Dv2, RealEstate10K, and Objaverse. \ourpe{} uses image features to predict per-patch transformation parameters. Best in \colorbox{besthl}{green}, second in \colorbox{sndhl}{blue}.}
\vspace{3pt}
\label{tab:main}
\resizebox{\textwidth}{!}{ 
\begin{tabular}{l ccc ccc ccc}
\toprule
& \multicolumn{3}{c}{CO3Dv2} & \multicolumn{3}{c}{RealEstate10K} & \multicolumn{3}{c}{Objaverse} \\
\cmidrule(lr){2-4}\cmidrule(lr){5-7}\cmidrule(lr){8-10}
Method & PSNR\,$\uparrow$ & SSIM\,$\uparrow$ & LPIPS\,$\downarrow$ & PSNR\,$\uparrow$ & SSIM\,$\uparrow$ & LPIPS\,$\downarrow$ & PSNR\,$\uparrow$ & SSIM\,$\uparrow$ & LPIPS\,$\downarrow$ \\
\midrule
LVSM~\citep{lvsm} & 18.617 & 0.601 & 0.618 & 25.282 & 0.802 & 0.130 & 29.676 & 0.923 & 0.098 \\
GTA~\citep{gta} & 19.482 & 0.612 & 0.573 & 26.076 & 0.824 & 0.110 & 30.500 & 0.929 & 0.079 \\
PRoPE~\citep{prope} & 19.495 & 0.626 & 0.583 & 26.142 & 0.826 & 0.111 & 30.588 & 0.929 & 0.078 \\
RayRoPE~\citep{rayrope} & 18.948 & 0.617 & 0.601 & 26.348 & 0.830 & 0.107 & 30.871 & 0.932 & 0.079 \\
URoPE~\citep{urope} & - & - & - & \snd{26.836} & \snd{0.838} & \snd{0.099} & 30.890 & 0.930 & 0.073 \\ \midrule
\ourpe{} (w. Learnable) & 19.612 & 0.630 & 0.549 & 26.419 & 0.831 & 0.105 & 30.879 & 0.932 & 0.071 \\
\ourpe{} (w. DINOv2) & \snd{19.732} & \snd{0.635} & \snd{0.519} & 26.701 & 0.833 & \snd{0.099} & \snd{31.365} & \snd{0.934} & \snd{0.060} \\
\ourpe{} (w. VGGT) & \best{20.025} & \best{0.647} & \best{0.462} & \best{27.410} & \best{0.848} & \best{0.091} & \best{31.622} & \best{0.937} & \best{0.057} \\
\bottomrule
\end{tabular}
}
\end{table*}

\vspace{3pt}
\begin{table*}[t]
\centering
\caption{Scaling on the DL3DV benchmark with a $53$M LVSM under $2$-, $4$-, and
$8$-view inputs. \ourpe{} consistently improves over PRoPE across all three input-view settings. Best in \colorbox{besthl}{green}.}
\label{tab:dl3dv}
\vspace{2pt}
\resizebox{\textwidth}{!}{
\begin{tabular}{l ccc ccc ccc}
\toprule
& \multicolumn{3}{c}{$2$ views} & \multicolumn{3}{c}{$4$ views} & \multicolumn{3}{c}{$8$ views} \\
\cmidrule(lr){2-4}\cmidrule(lr){5-7}\cmidrule(lr){8-10}
Method & PSNR\,$\uparrow$ & SSIM\,$\uparrow$ & LPIPS\,$\downarrow$
       & PSNR\,$\uparrow$ & SSIM\,$\uparrow$ & LPIPS\,$\downarrow$
       & PSNR\,$\uparrow$ & SSIM\,$\uparrow$ & LPIPS\,$\downarrow$ \\
\midrule
LVSM~\citep{lvsm}  & 17.718 & 0.459 & 0.474 & 19.331 & 0.571 & 0.317 & 20.441 & 0.615 & 0.273 \\
PRoPE~\citep{prope} & 18.574 & 0.503 & 0.459 & 19.957 & 0.602 & 0.289 & 20.653 & 0.621 & 0.271 \\
\ourpe{}         & \best{19.284} & \best{0.544} & \best{0.447} & \best{20.373} & \best{0.624} & \best{0.270} & \best{20.940} & \best{0.635} & \best{0.257} \\
\bottomrule
\end{tabular}
}
\end{table*}

\vspace{3pt}
\paragraph{Results.}

\begin{figure}[t]
\centering
\includegraphics[width=\linewidth]{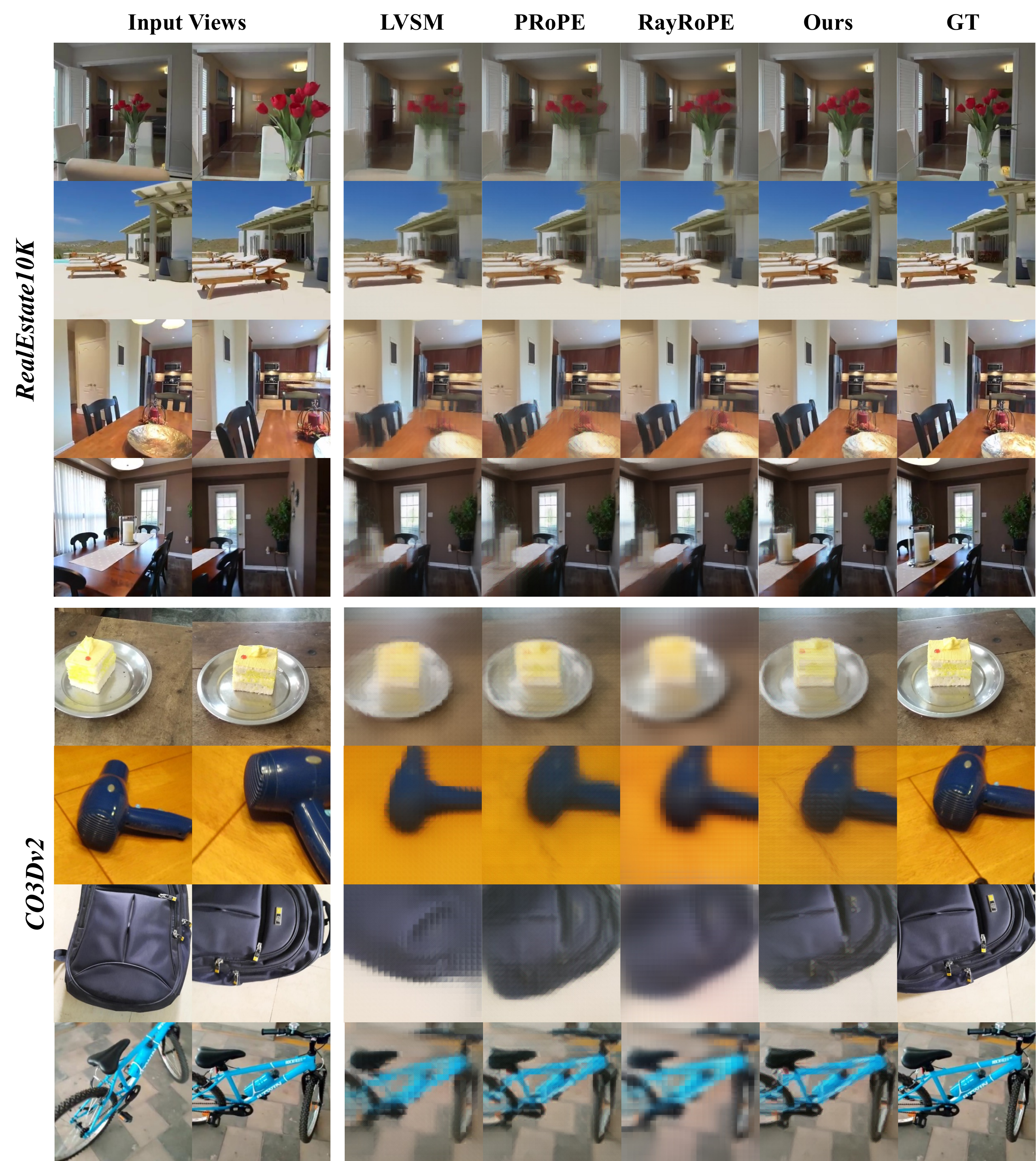}
\caption{\textbf{Qualitative novel-view synthesis on RealEstate10K and CO3Dv2.}
Each row shows one scene or object. From left to right, the columns contain two input
views, predictions from LVSM, PRoPE, RayRoPE, and \ourpe{}, and the
ground-truth target view (GT). }
\label{fig:qual}
\end{figure}

Table~\ref{tab:main} shows that the VGGT-based \ourpe{} outperforms baseline
PEs on CO3Dv2, RealEstate10K, and Objaverse. Relative to PRoPE, it gains
$0.53$, $1.27$, and $1.03$ dB in PSNR, respectively, with higher SSIM and
lower LPIPS. On DL3DV, \ourpe{} also outperforms PRoPE with $2$, $4$, and
$8$ reference views (Table~\ref{tab:dl3dv}). PSNR and SSIM gains are largest
with two views, suggesting greater benefits under sparse observations.
Qualitative comparisons on RealEstate10K show that \ourpe{} better preserves
object boundaries and fine details (Fig.~\ref{fig:qual}).

\subsection{Ablation Studies and Analysis}
\label{sec:ablation&analysis}


We investigate the design and behavior of \ourpe{} from three perspectives. First, we compare transformation parameterizations and feature sources for parameter prediction (Sec.~\ref{sec:ablation}). Second, we examine the contribution and quality of local geometry through a calibrated-camera comparison, qualitative visualizations, and an oracle geometry experiment (Sec.~\ref{sec:bridge}). 
Finally, we evaluate generalization under shifts in camera intrinsics and view sampling (Sec.~\ref{app:ood}).

\vspace{2pt}

\subsubsection{Parameterization and Feature Sources}
\label{sec:ablation}
\begin{figure}[t]
\centering
\includegraphics[width=\linewidth]{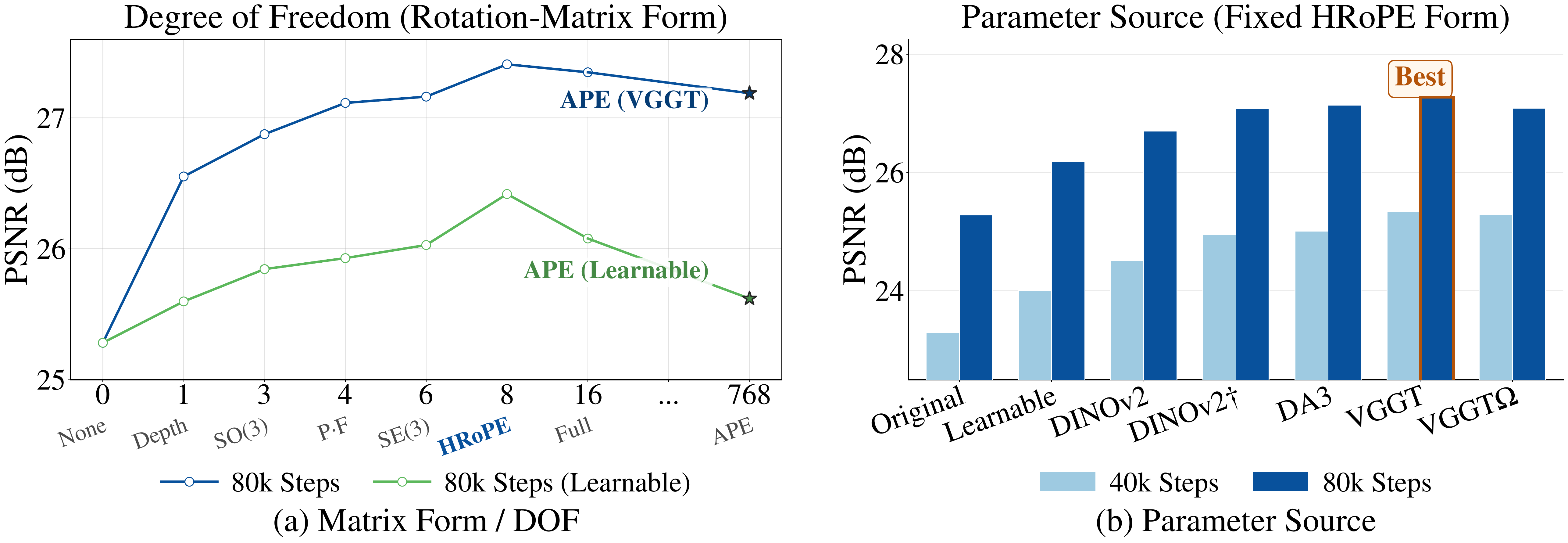}
\caption{\textbf{Ablations on RealEstate10K, measured by PSNR.} \textbf{(a)} Transformation parameterizations using frozen VGGT features (blue) or a predictor trained from scratch (green), evaluated after $80$k training steps. The x-axis labels denote the LVSM baseline (None), a depth term (Depth), a rotation ($\mathrm{SO}(3)$), the calibrated projection composed with a per-patch tangent-frame transform ($P\!\cdot\!F$), a rigid transformation ($\mathrm{SE}(3)$), our homography parameterization (HRoPE), and an unconstrained $4\times4$ matrix (Full). Stars denote the corresponding $768$-dimensional additive positional encodings (APE).
\textbf{(b)} Feature sources with the \ourpe{} parameterization
fixed, evaluated after $40$k and $80$k training steps.
DINOv2$^{\dagger}$ denotes the fine-tuned DINOv2 features from VGGT.}
\label{fig:ablation}
\end{figure}

\vspace{2pt}
\paragraph{Transformation Parameterization.}
Keeping the overall LVSM architecture unchanged, we compare
transformations with different geometric structures and degrees
of freedom: a depth term (``Depth''), a rotation ($\mathrm{SO}(3)$),
the calibrated $P\!\cdot\!F$ transformation, a rigid transformation
($\mathrm{SE}(3)$), the homography parameterization used by
\ourpe{}, and an unconstrained $4\times4$ matrix (``Full'').
Fig.~\ref{fig:ablation}(a) compares these variants against the
LVSM baseline (``None'').
Whether using a predictor trained from scratch or frozen VGGT
features, \ourpe{} achieves the highest PSNR among the evaluated
transformations. In both settings, it outperforms the unconstrained
$4\times4$ transformation despite using fewer degrees of freedom.
 This suggests that the homography structure provides an effective geometric inductive bias without requiring a fully unconstrained transformation. We additionally compare the homography encoding with a $768$-dimensional additive encoding constructed from the same VGGT features. The homography formulation performs better despite using the same feature source, indicating that the improvement is not solely due to access to pretrained VGGT features.

\vspace{2pt}

\paragraph{Parameter Source.}
We next fix the homography parameterization and vary the feature source used to predict its parameters. We compare a learnable predictor trained from scratch, including DINOv2~\citep{dinov2}, DepthAnything3~\citep{depthanything3}, VGGT~\citep{vggt}, and VGGT-$\Omega$~\citep{vggtomega}. Fig.~\ref{fig:ablation}(b) shows that the proposed parameterization improves over PRoPE even when its parameters are predicted from scratch or from DINOv2 features. Notably, DINOv2$^{\dagger}$ retains a 2D architecture, but its features are fine-tuned within VGGT's 3D pretraining pipeline and therefore acquire implicit 3D awareness, which helps explain their stronger performance than that of off-the-shelf DINOv2. Performance further improves with stronger pretrained feature extractors, with the geometry-aware VGGT backbone achieving the highest PSNR. These results show that the gains of \ourpe{} do not depend on a particular pretrained backbone, while more informative geometric features provide additional improvements.

\subsubsection{Analysis of Local Surface Geometry}
\label{sec:bridge}

\begin{table*}[t]
\centering
\vspace{3pt}
\caption{\textbf{Local surface geometry on RealEstate10K (Left)
\& Oracle geometry on ASE (Right).}
On RealEstate10K, PRoPE ($P\!\cdot\!F$) and \ourpe{} both use VGGT features for parameter prediction. 
On ASE, \ourpe{} uses either VGGT-predicted parameters or ground-truth camera and plane parameters.
Best and second-best results are highlighted in
\colorbox{besthl}{green} and \colorbox{sndhl}{blue}.}
\label{tab:bridge}

\small
\setlength{\tabcolsep}{3pt}
\renewcommand{\arraystretch}{1.15}

\begin{minipage}[t]{0.48\textwidth}
\centering
\begin{tabular*}{\linewidth}{@{\extracolsep{\fill}}lccc@{}}
\toprule
Method & PSNR$\uparrow$ & SSIM$\uparrow$ & LPIPS$\downarrow$ \\
\midrule
PRoPE
& 26.142 & 0.826 & 0.111 \\
PRoPE ($P\!\cdot\!F$)
& \snd{27.115} & \snd{0.840} & \snd{0.104} \\
\ourpe{}
& \best{27.410} & \best{0.848} & \best{0.091} \\
\bottomrule
\end{tabular*}
\end{minipage}
\hfill
\begin{minipage}[t]{0.48\textwidth}
\centering
\begin{tabular*}{\linewidth}{@{\extracolsep{\fill}}lccc@{}}
\toprule
Method & PSNR$\uparrow$ & SSIM$\uparrow$ & LPIPS$\downarrow$ \\
\midrule
PRoPE
& 27.829 & 0.802 & 0.264 \\
\ourpe{} (w. VGGT)
& \snd{28.555} & \snd{0.809} & \snd{0.198} \\
\ourpe{} (w. GT)
& \best{28.751} & \best{0.820} & \best{0.192} \\
\bottomrule
\end{tabular*}
\end{minipage}
\end{table*}

\paragraph{Local Geometry with Calibrated Cameras.}
To isolate the effect of local geometry, we start from PRoPE's GT camera projection $P$ and compose it with the per-patch tangent-frame transformation $F$ derived from predicted plane parameters, forming the calibrated $P\!\cdot\!F$ variant. We evaluate PRoPE, $P\!\cdot\!F$, and the fully predicted \ourpe{} on RealEstate10K using the setup in Sec.~\ref{sec:nvs}. As shown in Table~\ref{tab:bridge}, adding $F$ improves PRoPE from $26.142$ to $27.115$ in PSNR and from $0.826$ to $0.840$ in SSIM, while reducing LPIPS from $0.111$ to $0.104$. Because PRoPE and $P\!\cdot\!F$ share the same projection $P$, these gains support the contribution of the local surface transformation. \ourpe{} with predicted transformations further
improves all three metrics over the calibrated
$P\!\cdot\!F$ variant.

\begin{figure}[t]
\centering
\includegraphics[width=\linewidth]{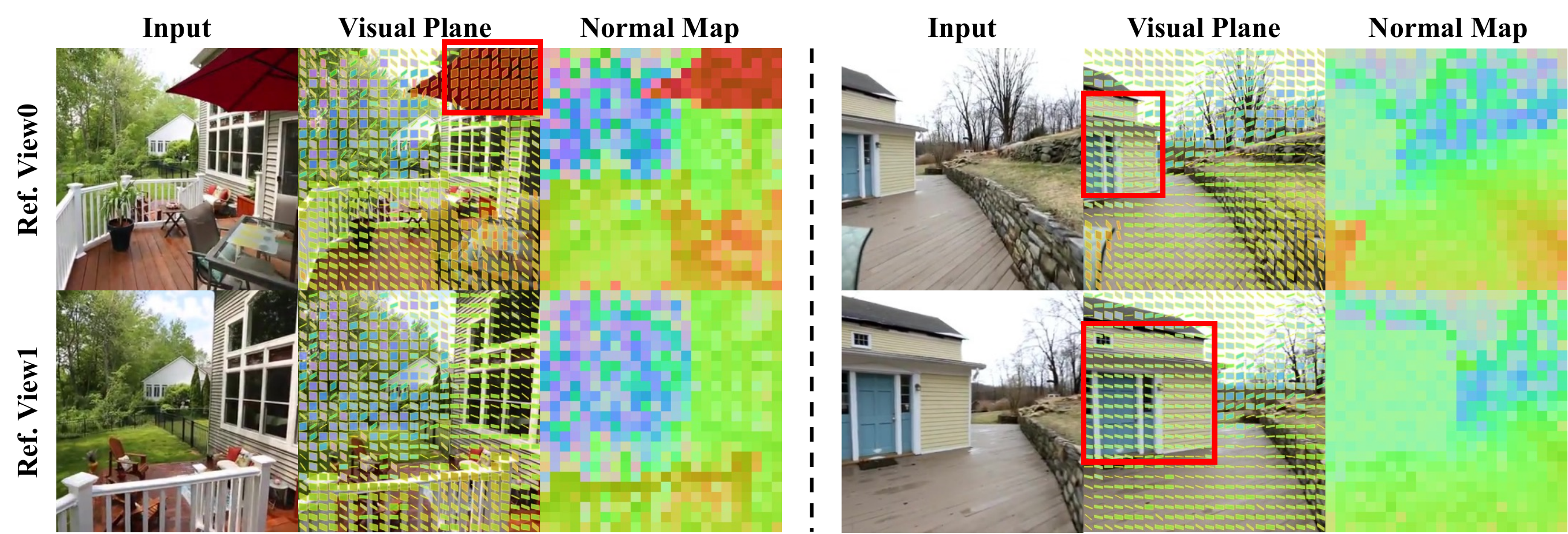}
\caption{\textbf{Predicted local planes on RealEstate10K.}
Two scenes are shown side by side; within each scene, the two rows correspond to the reference views. For each view, the columns show the input image, predicted
plane orientations overlaid on the image (Visual Plane), and per-patch normals
encoded as a color map (Normal Map). The red boxes mark the same approximately
planar surface across views, where the predicted orientations remain locally
coherent and cross-view consistent.}
\label{fig:planes}
\end{figure}

\paragraph{Predicted Local Geometry.}
Fig.~\ref{fig:planes} shows predicted orientations for the two input views using overlays and per-patch normal maps. Patches belonging to approximately planar regions exhibit similar predicted orientations, while corresponding regions across reference views show qualitatively consistent surface orientations when expressed in a common coordinate frame. The visualizations suggest coherent local structure without direct plane supervision. 

\paragraph{Oracle Geometry.}
Finally, we test whether more accurate, physically grounded parameters further improve \ourpe{}. We use Aria Synthetic Environments~\citep{scenescript}, which provides ground-truth depth, surface normals, and camera parameters, and construct the transformation $A$ in Eq.~\eqref{eq:A} using either VGGT-predicted parameters or ground-truth geometry. As shown in Table~\ref{tab:bridge}, replacing the VGGT predictions with ground-truth geometry improves PSNR from $28.555$ to $28.751$, while also improving SSIM and LPIPS. This result suggests that \ourpe{} benefits from more accurate geometric inputs under the evaluated oracle protocol.

\subsubsection{Out-of-Distribution Generalization}
\label{app:ood}

\begin{figure}[t]
    \centering
    \includegraphics[height=0.25 \linewidth]{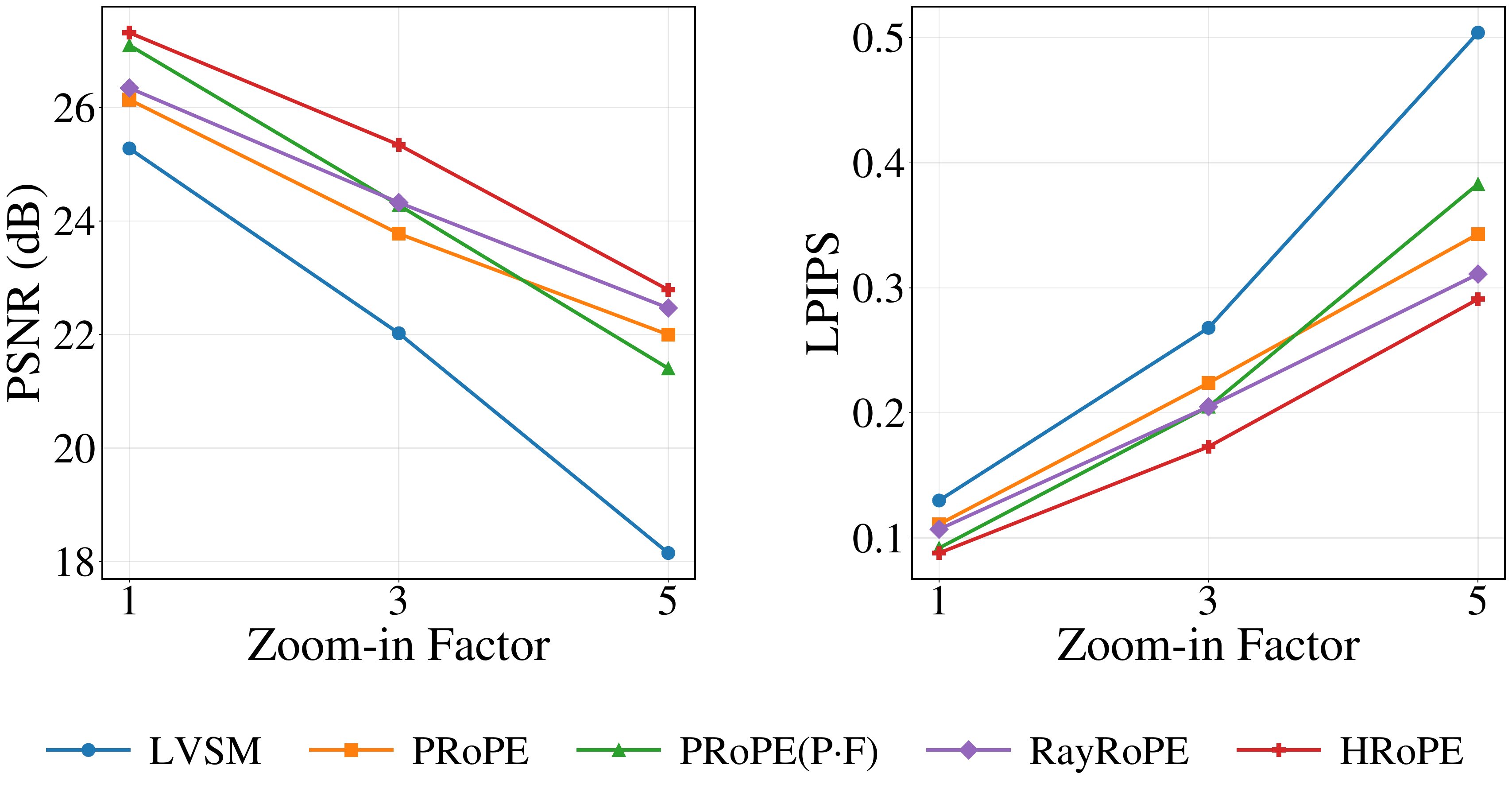}
    \hfill
    \includegraphics[height=0.25\linewidth]{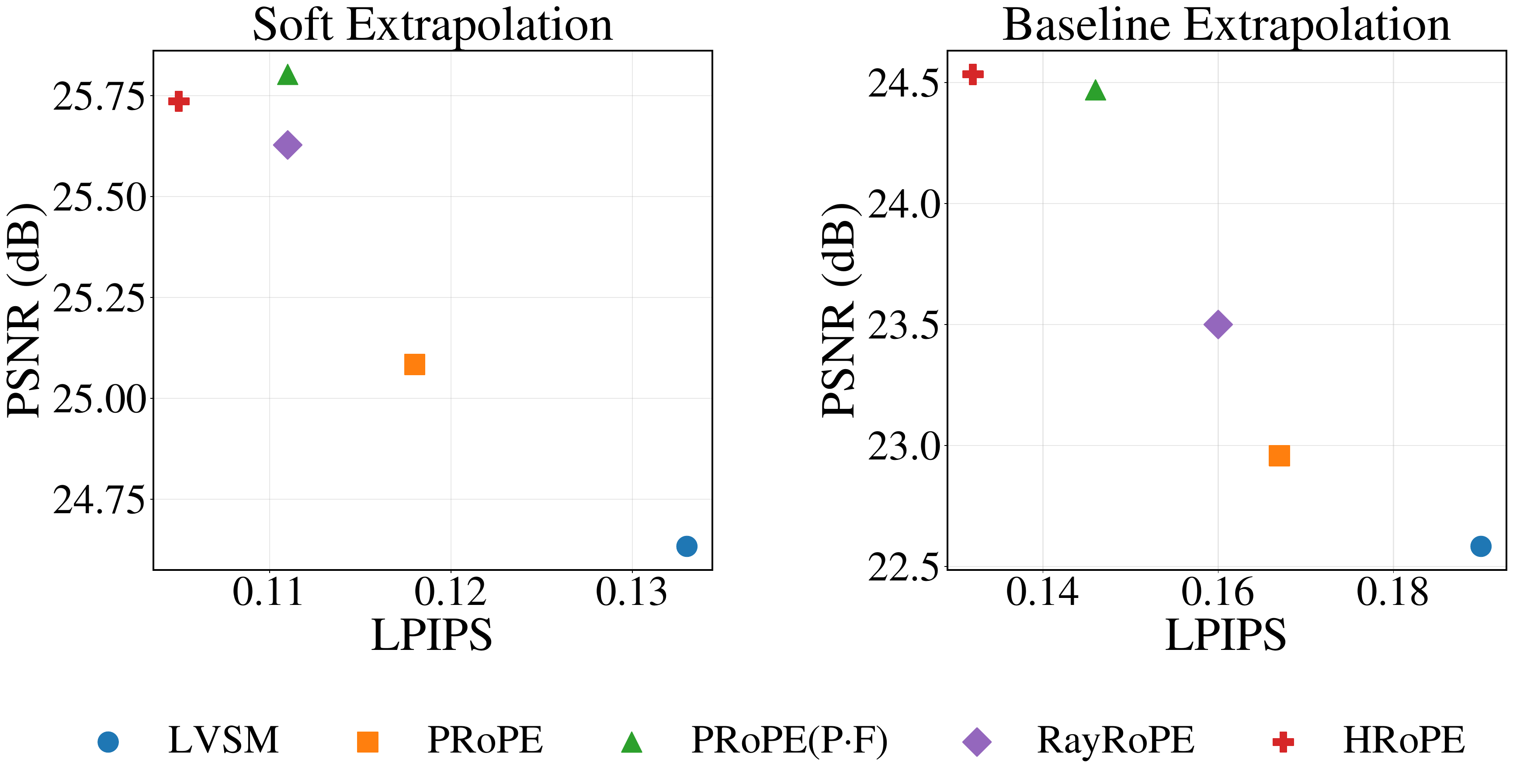}
    \caption{\textbf{Out-of-distribution generalization on RealEstate10K.}
    \textbf{Left pair:} PSNR ($\uparrow$) and LPIPS ($\downarrow$) as the unseen
    target-camera zoom factor increases from $1\times$ to $5\times$.
    \textbf{Right pair:} the PSNR--LPIPS trade-off under soft target-view
    extrapolation and an enlarged baseline between the two input views.}
    \label{fig:oodexp}
\end{figure}

\paragraph{Focal-length Variation.}
We first evaluate robustness to changes in camera intrinsics by varying the target-camera focal length at test time. Specifically, we scale the focal length from $1\times$ to $5\times$ its original value, where $1\times$ corresponds to the standard evaluation setting. As shown in Fig.~\ref{fig:oodexp}, \ourpe{} maintains strong performance as the zoom factor increases and achieves the lowest LPIPS across the evaluated focal lengths. In contrast, the calibrated $P\!\cdot\!F$ variant degrades more substantially under large zoom factors.

\paragraph{Target-view Extrapolation.}
We next extend the target-view sampling interval by $50\%$ at both ends, allowing target views to fall outside the interval bounded by the two reference views. Fig.~\ref{fig:oodexp} shows that \ourpe{} maintains its advantage over the baseline positional encodings as the target view moves beyond the training sampling regime.

\paragraph{Input-view Spacing.}
We additionally vary the temporal spacing between the two reference frames. We shift the input-frame interval from the training range of \([25,100]\) to \([75,150]\) at test time. \ourpe{} remains robust as the reference views become farther apart and achieves lower LPIPS across the evaluated intervals.

\section{Conclusion}

In this paper, we investigated local surface geometry as a prior for positional encoding
in multi-view transformers. Inspired by the slanted support windows of
PatchMatch Stereo, we introduced \ourpe{}, which uses the structure of
plane-induced homographies to parameterize per-patch transformations
and modulate attention through their relative composition.
The transformation parameters are predicted from patch features
and learned jointly with the view-synthesis model.
Experiments on RealEstate10K, CO3Dv2, Objaverse, and DL3DV demonstrate
improved novel view synthesis over the evaluated positional encodings.
Ablations support the contribution of the transformation structure,
with gains observed for predictors trained from scratch and further
improvements from pretrained features. A calibrated extension also
improves PRoPE by incorporating local planar frames while retaining
its camera projections. These findings support local surface geometry
as a useful inductive bias for multi-view attention.

\paragraph{Limitations.} \ourpe{} encourages correct cross-view correspondence through the structure of its
rectified matching, but does not guarantee it: because the plane parameters are predicted per patch and per view, the encoding offers no formal guarantee that the same physical surface is encoded identically across views, unlike methods that enforce such uniqueness by construction. In practice the learned planes are
cross-view consistent and the calibrated $P\!\cdot\!F$
bridge corroborates the underlying geometry, but a hard guarantee remains future
work. 





\newpage
\bibliography{iclr2027_conference}
\bibliographystyle{iclr2027_conference}

\clearpage

\begingroup
\renewcommand{\iclrruler}[1]{}
\hypersetup{hidelinks,linktoc=all}
\tableofcontents
\clearpage
\endgroup

\appendix

\section{Appendix}
\label{app:theory}

We first expand the notation and derivations behind the geometric positional
encodings summarized in Sec.~\ref{sec:prelim} (Appendix~\ref{app:gpe-background}).
Appendix~\ref{app:derivation} then derives the
plane-induced homography that computes a pixel in view $i$ directly from a pixel in
view $j$, and explains why a purely projective encoding cannot express this
correspondence. Appendix~\ref{app:matrix} then writes out in full the
block-diagonal operator that applies this homography across the head dimension.

\subsection{Background on Geometric Positional Encodings}
\label{app:gpe-background}

\paragraph{Notation.}
For a patch token $u$ in view $i$, let $\mathbf{q}_{i,u}$,
$\mathbf{k}_{i,u}$, and $\mathbf{v}_{i,u}$ denote its query, key, and value
vectors. The calibrated camera has intrinsics $K_i$ and a world-to-camera
transform
\begin{equation}
\label{eq:app-camera}
E_i=\begin{bmatrix}R_i&\mathbf{t}_i\\\mathbf{0}^{\top}&1\end{bmatrix},
\qquad \mathbf{X}_i=R_i\mathbf{X}+\mathbf{t}_i.
\end{equation}
Here, $R_i\in\mathrm{SO}(3)$ and $\mathbf{t}_i\in\mathbb{R}^3$ describe the
camera pose. These calibrated quantities are distinct from the patch-specific
parameters predicted by \ourpe{} in Sec.~\ref{sec:source}.

\paragraph{Rotary positional encoding.}
Let $\rho(u)$ denote RoPE's block-diagonal rotation for position $u$. Applying it
to a query at $u$ and a key at $v$ gives
\begin{equation}
\label{eq:app-rope}
\big(\rho(u)\mathbf{q}\big)^{\top}\big(\rho(v)\mathbf{k}\big)
=\mathbf{q}^{\top}\rho(u)^{\top}\rho(v)\mathbf{k}
=\mathbf{q}^{\top}\rho(v-u)\mathbf{k},
\end{equation}
so the positional dependence of the attention score is determined by the relative displacement $v-u$. The usual
$2$D extension applies this construction separately along the two image axes.

\paragraph{Camera-relative encodings.}
GTA~\citep{gta} and CaPE~\citep{cape} extend the same principle from image-grid
offsets to camera motion: their feature transformations make the interaction depend
on the relative extrinsic transform $E_iE_j^{-1}$, with method-specific
representations of its action on features. PRoPE~\citep{prope} additionally includes
intrinsics by lifting $K_i$ to homogeneous coordinates,
\begin{equation}
\label{eq:app-projective-factor}
\bar K_i=
\begin{bmatrix}
K_i & \mathbf{0}\\
\mathbf{0}^{\top} & 1
\end{bmatrix},
\qquad
P_i=\bar K_iE_i,
\qquad
P_iP_j^{-1}=\bar K_iE_iE_j^{-1}\bar K_j^{-1}.
\end{equation}
It applies $P_i^{\top}$ to queries and $P_j^{-1}$ to keys, yielding the relative
projective factor. This factor is invariant to the choice of
world frame: replacing every $E_i$ by $E_iT^{-1}$ for a shared rigid transform $T$
leaves $P_iT^{-1}(P_jT^{-1})^{-1}=P_iP_j^{-1}$ unchanged.


\subsection{Derivation of the Plane-Induced Homography}
\label{app:derivation}

Let view $j$ be the reference camera, with the world frame aligned to it, so a 3D point $\mathbf{X}_j\in\mathbb{R}^3$ (in camera-$j$ coordinates) projects to the homogeneous pixel $\tilde{\mathbf{x}}_j$ of view $j$ as
\begin{equation}
\label{eq:app-proj-a}
z_j\,\tilde{\mathbf{x}}_j = K_j\,\mathbf{X}_j
\quad\Longleftrightarrow\quad
\mathbf{X}_j = z_j\,K_j^{-1}\tilde{\mathbf{x}}_j,
\end{equation}
where $z_j>0$ is the (unknown) depth along the optical axis of camera-$j$. Let $(R_{i\leftarrow j},\mathbf{t}_{i\leftarrow j})$ be the rigid transform from the camera-$j$ frame to the camera-$i$ frame; the same point projects to view $i$ as
\begin{equation}
\label{eq:app-proj-b}
z_i\,\tilde{\mathbf{x}}_i = K_i\,(R_{i\leftarrow j}\,\mathbf{X}_j + \mathbf{t}_{i\leftarrow j}).
\end{equation}
Suppose that $\mathbf{X}_j$ lies on a local plane with unit normal $\mathbf{n}_j$ at orthogonal distance $d_j>0$ in the camera-$j$ frame,
\begin{equation}
\label{eq:app-plane}
\mathbf{n}_j^\top\mathbf{X}_j = d_j
\quad\Longleftrightarrow\quad
\tfrac{1}{d_j}\,\mathbf{n}_j^\top\mathbf{X}_j = 1.
\end{equation}
The plane assumption is what removes the unknown depth. Substituting $\mathbf{X}_j=z_jK_j^{-1}\tilde{\mathbf{x}}_j$ from Eq.~\eqref{eq:app-proj-a} into Eq.~\eqref{eq:app-plane} determines the depth at the ray–plane intersection,
\begin{equation}
\label{eq:app-depth}
z_j = \frac{d_j}{\mathbf{n}_j^\top K_j^{-1}\tilde{\mathbf{x}}_j}.
\end{equation}
The key algebraic step is to rewrite the translation in Eq.~\eqref{eq:app-proj-b} using the plane identity $\tfrac{1}{d_j}\mathbf{n}_j^\top\mathbf{X}_j=1$, so that $\mathbf{t}_{i\leftarrow j}=\mathbf{t}_{i\leftarrow j}\,(\tfrac{1}{d_j}\mathbf{n}_j^\top\mathbf{X}_j)$ and the affine map becomes linear in $\mathbf{X}_j$:
\begin{equation}
\label{eq:app-collapse}
R_{i\leftarrow j}\,\mathbf{X}_j + \mathbf{t}_{i\leftarrow j}
= R_{i\leftarrow j}\,\mathbf{X}_j + \mathbf{t}_{i\leftarrow j}\,\Big(\tfrac{1}{d_j}\mathbf{n}_j^\top\mathbf{X}_j\Big)
= \Big(\underbrace{R_{i\leftarrow j} + \tfrac{1}{d_j}\,\mathbf{t}_{i\leftarrow j}\,\mathbf{n}_j^\top}_{\displaystyle A}\Big)\mathbf{X}_j.
\end{equation}
Combining Eqs.~\eqref{eq:app-proj-a},
\eqref{eq:app-proj-b}, and \eqref{eq:app-collapse}
and eliminating the depth-dependent scale yields the plane-induced homography between the two pixels,
\begin{equation}
\label{eq:app-H}
\tilde{\mathbf{x}}_i \;\sim\; H_{i\leftarrow j}\,\tilde{\mathbf{x}}_j,
\qquad
H_{i\leftarrow j} \;=\; K_i\Big(R_{i\leftarrow j} + \tfrac{1}{d_j}\,\mathbf{t}_{i\leftarrow j}\,\mathbf{n}_j^\top\Big)K_j^{-1}
\;=\; K_i\,A\,K_j^{-1},
\end{equation}
Here, $\sim$ denotes equality up to a nonzero scale.
For points in front of both cameras, $z_i,z_j>0$,
and the scale $z_j/z_i$ is positive. The inner factor has the same rotation-plus-rank-one structure as
the parameterization in Eq.~\eqref{eq:A}, with $\kappa=1/d_j$.
This connection motivates the use of local plane parameters in
our learned per-patch transformations. (The minus-sign form $R_{i\leftarrow j}-\tfrac{1}{d_j}\mathbf{t}_{i\leftarrow j}\,\mathbf{n}_j^\top$ arises under the alternative plane convention $\mathbf{n}_j^\top\mathbf{X}_j+d_j=0$.)

Eq.~\eqref{eq:app-H} defines a pixel-to-pixel mapping conditioned on a local plane. In comparison, the relative projection $P_iP_j^{-1}$ used by PRoPE depends only on camera parameters and does not generally determine a unique correspondence between image pixels. A source pixel $\tilde{\mathbf{x}}_j$ specifies a viewing ray $\mathbf{X}_j=z_jK_j^{-1}\tilde{\mathbf{x}}_j$, but its depth $z_j$ remains unknown. For a nonzero camera baseline, this depth ambiguity generally constrains the corresponding pixel in view $i$ to an epipolar line. Given a plane hypothesis $(\mathbf{n}_j,d_j)$, Eq.~\eqref{eq:app-depth} determines the depth at the ray--plane intersection, provided that the intersection is finite and lies in front of the source camera. Substituting this depth into the projection equation yields the homography in Eq.~\eqref{eq:app-H}, which maps points on the plane between the two views. Unlike the camera-only relative projection used by PRoPE, a local plane resolves the depth ambiguity along a viewing ray and induces a pixel-to-pixel homography on that plane. This motivates HRoPE’s incorporation of local surface geometry into its per-patch transformations. The same principle underlies the slanted support windows in PatchMatch Stereo. 

\paragraph{Global-frame invariance.}
In the image-based instantiation of \ourpe{}, the per-patch
transforms $A_{i,u}$ are predicted from image features without
using externally specified world-frame extrinsics.
Changing the world coordinate frame re-expresses the cameras
and scene geometry but leaves the images and intrinsics unchanged.
Consequently, the predicted transforms and the relative factors
$H_{i,u}H_{j,v}^{-1}$ remain unchanged.
This establishes invariance of the predicted geometric encoding;
invariance of the full attention computation additionally requires
its input features to be independent of the world-frame choice.

For transforms constructed from explicit camera and plane geometry,
the scope is more restricted.
When both factors describe a shared physical plane, consistently
changing coordinates gives $H_i'=H_iG^{-1}$ and $H_j'=H_jG^{-1}$
for a common invertible matrix $G$, so
$H_i'(H_j')^{-1}=H_iH_j^{-1}$.
For different patch planes, the induced right multiplier can depend
on the plane, and this cancellation is not generally guaranteed.
Thus, the relative-product form alone does not establish
world-frame invariance for arbitrary geometry-derived transforms.
In our GT instantiation, both camera and plane parameters are
expressed in the first reference camera's coordinate frame
and normalized to a common scale, removing dependence on the
external world coordinate frame.


\subsection{The full block-diagonal encoding matrix}
\label{app:matrix}

Section~\ref{sec:homography} applies the encoding ``block-diagonally to the features''; here we write out that block-diagonal operator in full. Let the per-head feature dimension be $D$. For a token in view $i$ we split its $D$-dimensional query/key/value vector into three contiguous parts of sizes $D/2$, $D/4$, $D/4$. The first $D/2$ coordinates carry the geometric homography and the remaining $D/2$ carry a standard $2$D rotary encoding of the patch's pixel coordinates $(x,y)$, with $D/8$ frequencies for each axis. The block-diagonal encoding matrix for patch $(i,u)$ is
\begin{equation}
\label{eq:app-bigmat}
\mathcal{H}_{i,u} \;=\;
\underbrace{H_{i,u}^{\oplus D/6}}_{\text{geometry: } D/2 \text{ dims}}
\;\oplus\;
\underbrace{\Phi_x}_{D/4}\;\oplus\;\underbrace{\Phi_y}_{D/4}
\;=\;
\begin{bmatrix}
H_{i,u} & & & & \\
 & \ddots & & & \\
 & & H_{i,u} & & \\
 & & & \Phi_x & \\
 & & & & \Phi_y
\end{bmatrix},
\end{equation}
Here, $\oplus$ denotes a direct sum and $H^{\oplus r}$ contains $r$ copies of the $3\times3$ matrix $H$.  The geometric block is $D/6$ copies of the single $3\times3$ matrix $H_{i,u}$ (here $D/6=8$), because each copy acts on a $3$-vector and $3\cdot(D/6)=D/2$; this is what makes ``apply $H$ to the feature'' well-defined when $D/2 > 3$. The two rotary blocks are the usual $2$D-RoPE rotations
\begin{equation}
\Phi_x = \bigoplus_{\ell=1}^{D/8}
\begin{bmatrix}
\cos(x\,\omega_\ell) & -\sin(x\,\omega_\ell)\\[2pt]
\sin(x\,\omega_\ell) & \cos(x\,\omega_\ell)
\end{bmatrix},
\qquad
\omega_\ell = \theta^{-2\ell/(D/2)},
\end{equation}
and $\Phi_y$ is identical with the patch row coordinate $y$; together they encode the in-image patch position, complementing the cross-view geometry in the first block. Attention uses $\mathcal{H}_{i,u}^{\top}$ on the query and $\mathcal{H}_{j,v}^{-1}$ on the key, so that the geometric blocks contract to $H_{i,u}H_{j,v}^{-1}$ (Eq.~\eqref{eq:logit}) and the rotary blocks contract to a relative rotation in $(x,y)$, recovering the standard $2$D-RoPE relative property. For target-view tokens, which carry no predicted plane, the
geometric block reduces to $K_i$ alone (i.e.\ $A_{i,u}=I$),
encoding only camera intrinsics while retaining the 2D
rotary blocks.

\subsection{Integrating \ourpe{} into multi-view attention}
\label{app:integration}

\begin{figure}[!ht]
    \centering
    \includegraphics[width=0.6\linewidth]{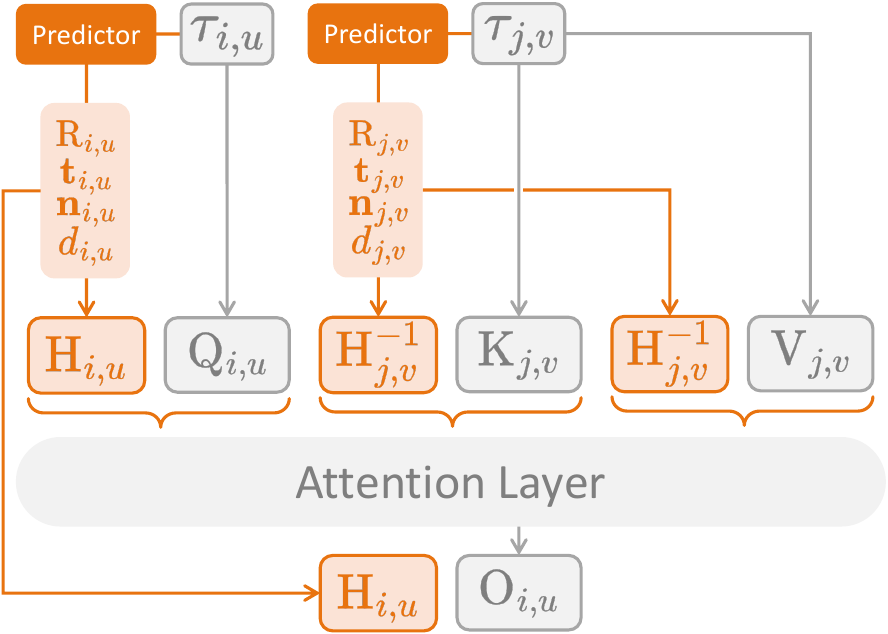}
\caption{\textbf{Applying \ourpe{} to Multi-view Attention.}
    Here, $\tau_{i,u}$ denotes the token corresponding to image patch $u$
    in view $i$. A predictor maps each patch token $\tau_{i,u}$ to the
    transformation parameters
$(R_{i,u},\mathbf{t}_{i,u},\mathbf{n}_{i,u},\kappa_{i,u})$,
    from which $H_{i,u}=K_iA_{i,u}$ is constructed. Queries are transformed
    by $\mathcal{H}_{i,u}^{\top}$, while keys and values are transformed by
    $\mathcal{H}_{j,v}^{-1}$. The aggregated value is subsequently mapped
    to the query token's frame by $\mathcal{H}_{i,u}$. Only one query--key
    pair is shown for clarity.}
    \label{fig:hrope-attention}
\end{figure}

\ourpe{} modifies multi-view self-attention through token-specific linear
operators while leaving the other transformer components unchanged. For
$N$ views with $P$ patches each, the layer receives $NP$ tokens, per-patch
transforms $\{A_{i,u}\}$, and camera intrinsics $\{K_i\}$. We construct
$\mathcal{H}_{i,u}$ as in Eq.~\eqref{eq:app-bigmat} and apply
$\mathcal{H}_{i,u}^{\top}$ to each query and
$\mathcal{H}_{j,v}^{-1}$ to each key and value. The resulting attention
logit for tokens $(i,u)$ and $(j,v)$ contains the relative operator
$\mathcal{H}_{i,u}\mathcal{H}_{j,v}^{-1}$. After attention aggregation,
the output is transformed by $\mathcal{H}_{i,u}$, expressing the
aggregated features in the query token's coordinate system. The inverse
of $A_{i,u}=R_{i,u}+\kappa_{i,u}\mathbf{t}_{i,u}
\mathbf{n}_{i,u}^{\top}$ is computed in closed form using
Eq.~\eqref{eq:sm}. Fig.~\ref{fig:hrope-attention} illustrates these
operations, and Algorithm~\ref{alg:hope} specifies the forward pass.

\begin{algorithm}[t]
\caption{\ourpe{} multi-view self-attention layer (single head).}
\label{alg:hope}
\begin{algorithmic}[1]
\Require queries/keys/values $Q,K,V\in\mathbb{R}^{(NP)\times D}$; per-patch matrices $\{A_{i,u}\}$, intrinsics $\{K_i\}$; RoPE coeffs $(\Phi_x,\Phi_y)$
\Statex \textbf{Notation:} $\mathcal{H}_{i,u}$ is the block-diagonal operator of Eq.~\eqref{eq:app-bigmat} built from the per-patch $3{\times}3$ matrix $H_{i,u}$ and $(\Phi_x,\Phi_y)$. $\mathrm{BD}(\mathcal{H},f)$ applies the corresponding per-token operators to the features $f$.
\For{each token $u$ in view $i$} \Comment{build the per-token geometric factor}
  \If{$i$ is an input view} $H_{i,u} \gets K_i\,A_{i,u}$
  \Else{} $\;H_{i,u} \gets K_i$ \Comment{target view: projective fallback, $A_{i,u}=I$}
  \EndIf
\EndFor
\State $\tilde{Q} \gets \mathrm{BD}\!\big(\mathcal{H}^{\top},\,Q\big)$ \label{line:qstart} \Comment{query side: $H_{i,u}^{\top}$}
\State $\tilde{K} \gets \mathrm{BD}\!\big(\mathcal{H}^{-1},\,K\big)$ \Comment{key side: $H_{i,u}^{-1}$ via Sherman--Morrison}
\State $\tilde{V} \gets \mathrm{BD}\!\big(\mathcal{H}^{-1},\,V\big)$ \label{line:qend}
\State $S \gets \tilde{Q}\tilde{K}^{\top}/\sqrt{D}$
  \Comment{logit $=\mathbf{q}^{\top}H_{i,u}H_{j,v}^{-1}\mathbf{k}$, Eq.~\eqref{eq:logit}}
\State $O \gets \mathrm{softmax}(S)\,\tilde{V}$
\State $O \gets \mathrm{BD}\!\big(\mathcal{H},\,O\big)$ \label{line:out} \Comment{back to query frame}
\State \Return $O$
\end{algorithmic}
\end{algorithm}


\subsection{Additional experiment details}
\label{app:details}

\paragraph{Datasets.}
We use five datasets in our experiments: RealEstate10K, CO3Dv2,
Objaverse, DL3DV, and Aria Synthetic Environments (ASE).
We describe the data splits, preprocessing, and frame sampling
for each dataset below.

\begin{itemize}[leftmargin=1.4em,itemsep=2pt,topsep=2pt]
\item \textbf{RealEstate10K}~\citep{real10k}: frames are extracted from the source
  YouTube videos at the official timestamps, with the given per-sequence intrinsics
  and per-frame world-to-camera poses converted into the common format above. During
  training we sample input/target frames with an index gap in $[25,100]$.
\item \textbf{CO3Dv2}~\citep{co3d}: we select a higher-quality subset and split it
  into a $31$k-scene training set and a $1$k-scene test set, both covering all $52$
  categories uniformly. We use a frame gap in $[25,50]$ during training.
\item \textbf{Objaverse}~\citep{deitke2023objaverse}: from the G-buffer Objaverse
  set~\citep{zuo2024sparse3d} we keep objects with fewer than $20$k faces ($96$k objects) and render $30$
  views each at $336$ resolution in Blender. The camera looks down at the object from
  a fixed pitch of about $45^\circ$, at radius $2.0$ and field of view $40^\circ$; for
  each object we sample views uniformly within a randomly placed $180^\circ$ azimuth
  window and with random perturbations to azimuth and elevation. We render on a transparent background composited onto white, and
  sample views with an angular gap in $[30^\circ,60^\circ]$ during training.
\item \textbf{DL3DV}~\citep{dl3dv}: we discard portrait-orientation videos and train on all remaining $10$k scenes. For
  evaluation we use the RayZer benchmark (the $10$k scenes not included in the
  benchmark are used for training)~\citep{rayzer}. Training frame gaps follow the input regime: $[25,50]$ for the $2$- and $4$-view models and $[25,100]$ for the $8$-view model.
\item \textbf{Aria Synthetic Environments}~\citep{scenescript}: we split the scenes at random into a $95$k-scene training set and
  a $5$k-scene test set, and use a frame gap in $[8,12]$ during training. This
  synthetic dataset renders a depth map for every frame, from which we derive the
  ground-truth plane parameters used in Sec.~\ref{sec:bridge}. Since a patch spans
  many pixels, we fit one plane per $14\times14$ patch using least squares on its
  inverse-depth samples. We express the fitted plane as
  $\mathbf{n}_c^\top\mathbf{X}_c=d_c$ in the camera frame and choose $d_c>0$ so that
  normals are oriented consistently across views. Patches that are more than half
  invalid, or whose median relative residual exceeds $5\%$ and are therefore poorly
  described by a single plane, fall back to a fronto-parallel plane at their median
  depth.
\end{itemize}

\paragraph{Training configuration.}
We train every model on four NVIDIA RTX 4090 ($48$\,GB) GPUs with a per-GPU batch
size of $16$, using a cosine learning-rate schedule with peak rate $4\times10^{-4}$
and a warmup of $2500$ iterations. All scene-level models are trained for $80$k
iterations at a fixed resolution of $336\times336$ obtained by center cropping. Camera
poses are normalized following the convention that matches the input regime. With few
input views (e.g.\ $2$ views) we follow PRoPE~\citep{prope}: the first reference
camera is set to the identity (its frame becomes the world origin) and all
translations are scaled by the distance between the first and second reference
cameras, so that this distance is unit length. With many input views we follow the
original LVSM~\citep{lvsm} convention: the scene is translated and rotated to align
with the mean camera center and mean viewing direction, then rescaled by the maximum
absolute camera-center coordinate times a fixed factor of $1.35$ to a common scale.
Our objective combines a photometric term with a perceptual loss; we weight the
perceptual term at $0.5$ for scene datasets and lower it to $0.1$ for the object
dataset.

\paragraph{Numerical stability of the predicted transformation.}
\label{app:clamp}
To stabilize training, we clamp the predicted translation $\mathbf{t}$ and inverse
plane distance $\kappa$ to $[-0.5,0.5]$, normalize $\mathbf{n}$, and construct
$\mathrm{R}$ to be orthonormal. This bounds the rank-one update in
$\mathrm{A}=\mathrm{R}+\kappa\mathbf{t}\mathbf{n}^{\top}$ and guarantees
$\det\mathrm{A}\ge 1-\sqrt{3}/4\approx0.57$. We additionally clamp the
Sherman--Morrison denominator to $0.1$ and the entries of $\mathrm{A}$ and
$\mathrm{A}^{-1}$ to $[-5,5]$. Both matrices are computed in \texttt{float32}
and then cast to the training precision.

\paragraph{Training cost.}
Table~\ref{tab:traincost} reports the final RealEstate10K PSNR and training cost
of the parameter-source variants, all trained for $80$k iterations. At the same
cost of $33$ GPU-hours, the learned predictor improves PSNR from $25.282$ to
$26.419$, showing that the gain does not depend solely on an expensive pretrained
source. Among the pretrained sources, DA3 achieves $27.137$ PSNR in $48$
GPU-hours, compared with $27.410$ PSNR in $76$ GPU-hours for VGGT.

\begin{table}[t]
\centering
\caption{Training cost of the parameter-source variants
(Sec.~\ref{sec:ablation}), all under the same $80$k-iteration schedule; PSNR is the
final RealEstate10K score. DINOv2$^\dagger$ is the fine-tuned DINOv2 taken from inside
VGGT. Best in \textbf{bold}, second \uline{underlined}.}
\label{tab:traincost}
\vspace{2pt}
\begin{tabular}{lcc}
\toprule
Source & PSNR\,$\uparrow$ & GPU-hours \\
\midrule
Original (LVSM)        & $25.282$ & $33$ \\
Learnable              & $26.419$ & $33$ \\
DINOv2~\citep{dinov2}                 & $26.701$ & $46$ \\
DINOv2$^\dagger$~\citep{dinov2}       & $27.085$ & $72$ \\
DA3~\citep{depthanything3}                    & \uline{$27.137$} & $48$ \\
VGGT~\citep{vggt}                   & \textbf{27.410} & $76$ \\
VGGT-$\Omega$~\citep{vggtomega}           & $27.090$ & $64$ \\
\bottomrule
\end{tabular}
\end{table}

\subsection{Generalization Beyond Architectures}
\label{app:transfer}

\ourpe{} can be integrated into multi-view attention by transforming its queries,
keys, and values. We evaluate this design in two additional architectures:
a generative multi-view diffusion model and a spatial multimodal large language
model.

\paragraph{Multi-view Diffusion Model.}
As a first step outside that setting, we transfer \ourpe{} to a generative architecture
with a different decoder and a different native encoding. We test on EscherNet~\citep{cape}, a large
($\sim\!800$M-parameter) multi-view diffusion model whose native encoding is the
camera positional encoding CaPE, by replacing CaPE with \ourpe{} while leaving the rest of
the model and its training recipe untouched. We train on
Objaverse~\citep{deitke2023objaverse} ($96$k objects) following EscherNet's setting,
feeding $3$ reference views during training, and evaluate the trained models with $1$,
$5$, and $10$ input views to probe both sparse- and denser-view regimes.

Table~\ref{tab:eschernet} shows that \ourpe{} improves EscherNet on every metric at every
view count, supporting the applicability of the encoding beyond LVSM and feed-forward
decoding.

\begin{table}[t]
\centering
\caption{Transferring \ourpe{} to EscherNet~\citep{cape}, an $\sim\!800$M-parameter
multi-view diffusion model, on Objaverse. We replace EscherNet's native CaPE encoding
with \ourpe{} and evaluate with $1$, $5$, and $10$ input views. Best in \colorbox{besthl}{green}.}
\label{tab:eschernet}
\resizebox{\textwidth}{!}{
\begin{tabular}{l ccc ccc ccc}
\toprule
& \multicolumn{3}{c}{$1$ view} & \multicolumn{3}{c}{$5$ views} & \multicolumn{3}{c}{$10$ views} \\
\cmidrule(lr){2-4}\cmidrule(lr){5-7}\cmidrule(lr){8-10}
Method & PSNR\,$\uparrow$ & SSIM\,$\uparrow$ & LPIPS\,$\downarrow$
       & PSNR\,$\uparrow$ & SSIM\,$\uparrow$ & LPIPS\,$\downarrow$
       & PSNR\,$\uparrow$ & SSIM\,$\uparrow$ & LPIPS\,$\downarrow$ \\
\midrule
EscherNet + CaPE & 17.689 & 0.858 & 0.181 & 22.065 & 0.899 & 0.114 & 23.666 & 0.914 & 0.089 \\
EscherNet + \ourpe{} & \best{19.900} & \best{0.883} & \best{0.135} & \best{23.075} & \best{0.911} & \best{0.092} & \best{24.515} & \best{0.923} & \best{0.074} \\
\bottomrule
\end{tabular}
}
\end{table}

\begin{table}[t]
\centering
\caption{Transferring \ourpe{} to a video multimodal LLM, evaluated on
VSI-Bench~\citep{vsibench}. Our model uses LLaVA-Video-7B-Qwen2 as the backbone with
\ourpe{} applied to the visual tokens, fine-tuned on the VSI-Bench training split at $32$
frames;  the preceding
row reports a fine-tuned 7B video model without any geometric encoding~\citep{vlm3r}. Frame budgets for the remaining methods are given in the text.
Rank is taken over the overall average. Best in \colorbox{besthl}{green}.}
\label{tab:vsibench}
\setlength\tabcolsep{3pt}
\resizebox{\textwidth}{!}{
\begin{tabular}{l|cccc|cccc|cc}
\toprule
& \multicolumn{4}{c|}{Numerical question} & \multicolumn{4}{c|}{Multiple-choice question} & & \\
\cmidrule(lr){2-5}\cmidrule(lr){6-9}
Method & Obj.\ Cnt. & Abs.\ Dist. & Obj.\ Size & Room Size
       & Rel.\ Dist. & Rel.\ Dir. & Route Plan & Appr.\ Order & Avg.\,$\uparrow$ & Rank \\
\midrule
\multicolumn{1}{l|}{\textit{Proprietary models}} & & & & & & & & & & \\
GPT-4o~\citep{gpt4o}                 & 46.2 & 5.3  & 43.8 & 38.2 & 37.0 & 41.3 & 31.5 & 28.5 & 34.0 & 9 \\
Gemini-1.5 Pro~\citep{gemini15}      & 56.2 & 30.9 & 64.1 & 43.6 & 51.3 & 46.3 & 36.0 & 34.6 & 45.4 & 4 \\
\midrule
\multicolumn{1}{l|}{\textit{Open-source models}} & & & & & & & & & & \\
LongVILA-8B~\citep{longvila}         & 29.1 & 9.1  & 16.7 & 0.0  & 29.6 & 30.7 & 32.5 & 25.5 & 21.6 & 13 \\
LongVA-7B~\citep{longva}             & 38.0 & 16.6 & 38.9 & 22.2 & 33.1 & 43.3 & 25.4 & 15.7 & 29.2 & 12 \\
VILA-1.5-40B~\citep{vila15}          & 22.4 & 24.8 & 48.7 & 22.7 & 40.5 & 25.7 & 31.5 & 32.9 & 31.2 & 11 \\
Qwen2.5VL-7B~\citep{qwen25vl}        & 40.9 & 14.8 & 43.4 & 10.7 & 38.6 & 38.5 & 33.0 & 29.8 & 33.0 & 10 \\
InternVL2-40B~\citep{internvl2}      & 34.9 & 26.9 & 46.5 & 31.8 & 42.1 & 32.2 & 34.0 & 39.6 & 36.0 & 8 \\
Qwen2.5VL-72B~\citep{qwen25vl}       & 25.1 & 29.3 & 54.5 & 38.8 & 38.2 & 37.0 & 34.0 & 28.9 & 37.0 & 7 \\
LLaVA-OneVision-72B~\citep{llavaov}  & 43.5 & 23.9 & 57.6 & 37.5 & 42.5 & 39.9 & 32.5 & 44.6 & 40.2 & 6 \\
LLaVA-Video-72B~\citep{llavavideo}   & 48.9 & 22.8 & 57.4 & 35.3 & 42.4 & 36.7 & 35.0 & \best{48.6} & 40.9 & 5 \\
Spatial-MLLM-4B~\citep{spatialmllm}  & 65.3 & 34.8 & 63.1 & 45.1 & 41.3 & 46.2 & 33.5 & 46.3 & 48.4 & 3 \\
\midrule
LLaVA-Video-7B (fine-tuned)~\citep{vlm3r}
                                     & 70.6 & 43.6 & \best{70.8} & 63.7 & \best{64.9} & 68.9 & 40.7 & 38.5 & 57.7 & 2 \\
LLaVA-Video-7B + \ourpe{} (ours)     & \best{70.9} & \best{51.6} & 69.8 & \best{65.9} & 64.1 & \best{76.0} & \best{46.4} & 31.6 & \best{59.5} & \best{1} \\
\bottomrule
\end{tabular}
}
\end{table}

\paragraph{Spatial Reasoning in Multimodal LLMs.}
We further evaluate how \ourpe{} enhances the spatial reasoning ability of multimodal LLMs. Visual tokens in these models are indexed only by their 2D raster location within a frame and their ordinal position in the frame sequence, so self-attention has no explicit representation of the 3D geometric relationship between observations of the same surface. Our proposed \ourpe{} enriches the visual tokens of a pretrained MLLM with
surface-aware positional information and cross-view correspondence cues, while modifying only their positional encoding. Concretely, we build on LLaVA-Video-7B~\citep{llavavideo}, sampling $32$ frames per video and processing them with a frozen VGGT~\citep{vggt} aggregator in parallel with the model's native vision tower. From VGGT's per-patch features, the prediction head described in Sec.~\ref{sec:source} regresses $(R_{i,u},\mathbf{t}_{i,u},\mathbf{n}_{i,u},\kappa_{i,u})$, which parameterize the per-patch matrix $A_{i,u}$ in Eq.~\eqref{eq:A}. We then insert the block-diagonal operator in Eq.~\eqref{eq:app-bigmat} into the language model's self-attention, applying it only to visual tokens in layers $L\!-\!10$ through $L\!-\!2$ of the $28$-layer backbone ($L=28$): the geometric halves of each head's query and key for token $(i,u)$ are transformed by $H_{i,u}^{\top}$ and $H_{i,u}^{-1}$, respectively, where $H_{i,u}=K_iA_{i,u}$, while the remaining halves retain the native rotary encoding. In this way, the visual stream of the MLLM is explicitly conditioned on per-patch surface geometry, while text tokens and the temporal ordering of frames are left unchanged, making \ourpe{} a strictly additive augmentation to the pretrained VLM. Following VLM-3R~\citep{vlm3r}, we fine-tune the model for 3 epochs on its released VSI-Bench training split, which combines spatial question--answer pairs over ScanNet~\citep{dai2017scannet} scenes with embodied route-planning instances, while keeping VGGT frozen throughout.

We report the evaluation results on VSI-Bench~\citep{vsibench} in Table~\ref{tab:vsibench}. Our model and the fine-tuned control take $32$ frames per video as input. For Spatial-MLLM~\citep{spatialmllm} and the Qwen2.5VL series~\citep{qwen25vl}, we use $16$ frames as input. For the other open-source methods and GPT-4o~\citep{gpt4o}, we follow the VSI-Bench evaluation protocol to select the number of input frames (ranging from $8$ to $32$ frames). For Gemini-1.5 Pro~\citep{gemini15}, it samples video frames at $1$ FPS. \ourpe{} reaches an average of $59.5$, ahead of every model in the table and $1.8$ points above the same backbone fine-tuned on the identical data without it. The gains concentrate in the categories that require relating a surface across frames, including absolute distance ($+8.0$), relative direction ($+7.1$), and route planning ($+5.7$). Categories answerable from a single frame show smaller changes. This pattern is consistent with the encoding's emphasis on local surface geometry. The exception is appearance order, which falls $6.9$ points: it is the one purely temporal task in the suite, and \ourpe{} spends half of each visual token's head dimension on the plane-induced map, weakening the ordinal cue it depends on. Since the language backbone, the fine-tuning data, and the frame budget are held fixed across the two rows, both the improvements and this regression are attributable solely to the positional encoding. This controlled comparison substantiates our central claim: the plane-induced homography benefits self-attention whenever it must establish correspondence between two observations of the same surface, irrespective of whether those observations are rendered, denoised, or sampled from video.


\subsection{Additional Qualitative Results}
\label{app:morevis}

\begin{figure}[t]
\centering
\includegraphics[width=\linewidth]{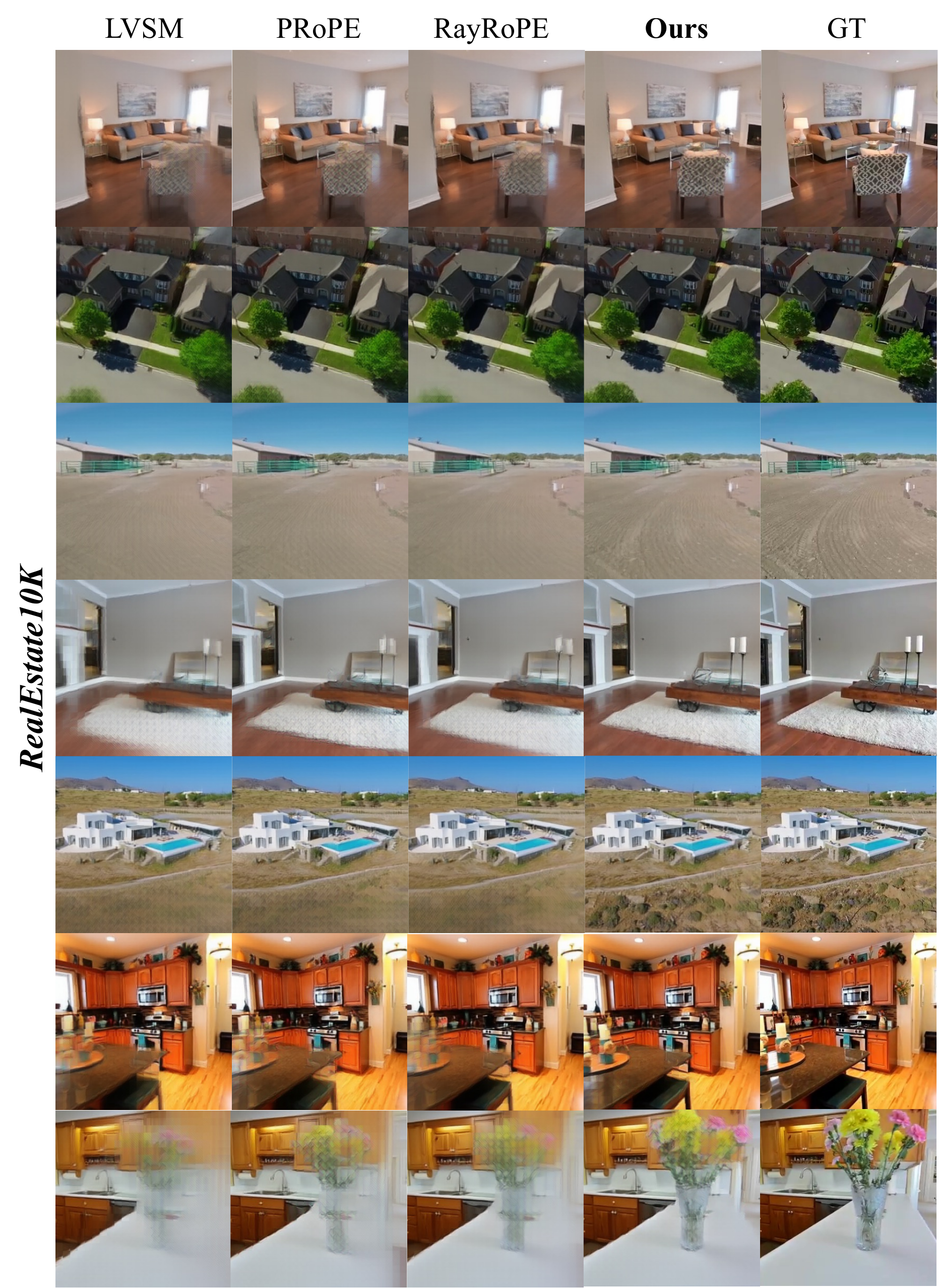}
\caption{\textbf{Additional qualitative comparison on RealEstate10K.}
Each row is one scene. From left to right, the columns show novel views synthesized
by LVSM, PRoPE, RayRoPE, and \ourpe{} (Ours), followed by the ground-truth target
view (GT). All methods use the same backbone and inputs and differ only in their
positional encoding.}
\label{fig:more-re10k}
\end{figure}

\begin{figure}[t]
\centering
\includegraphics[width=\linewidth]{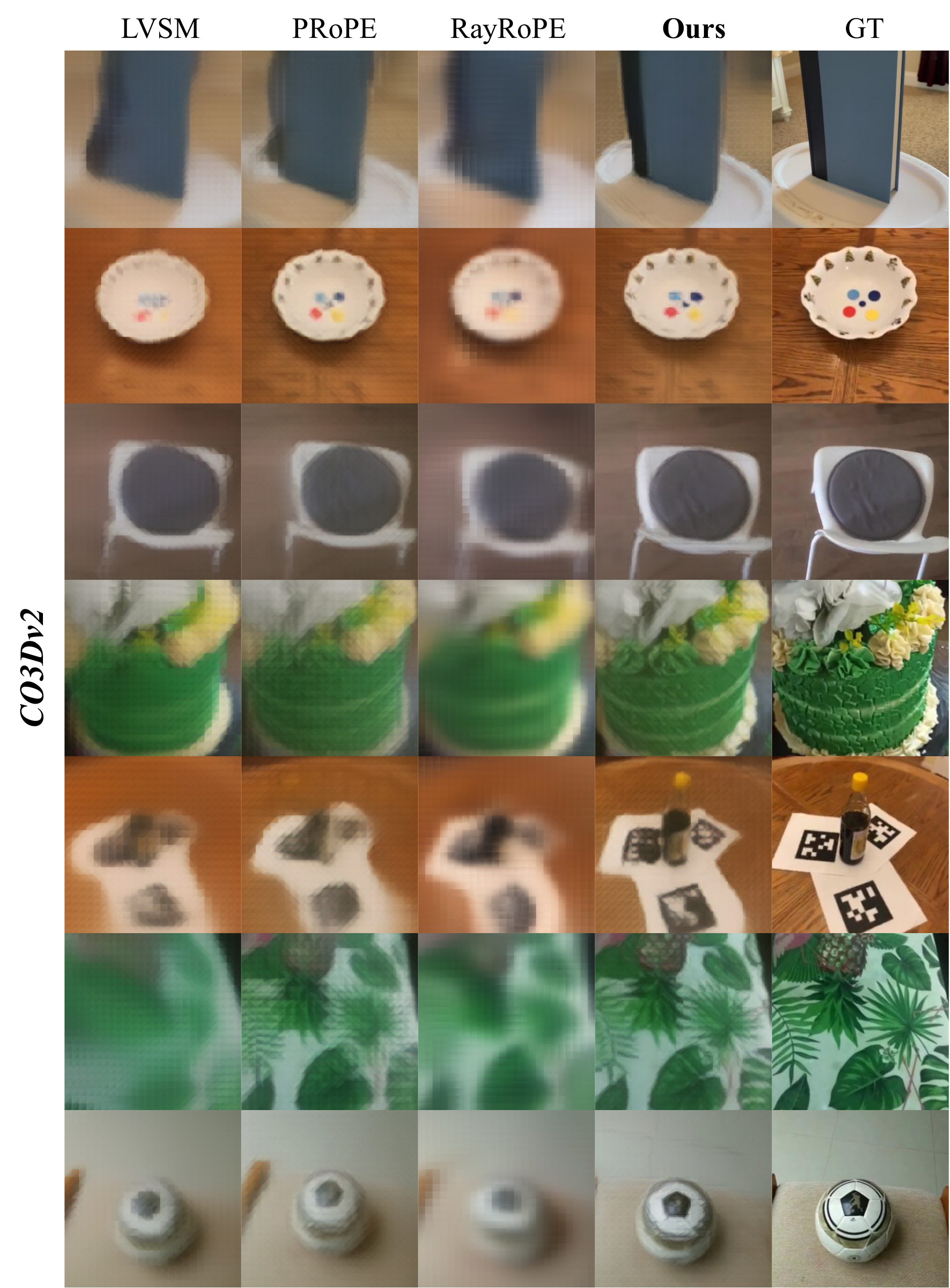}
\caption{\textbf{Additional qualitative comparison on CO3Dv2.}
Each row is one object. From left to right, the columns show novel views synthesized
by LVSM, PRoPE, RayRoPE, and \ourpe{} (Ours), followed by the ground-truth target
view (GT). All methods use the same backbone and inputs and differ only in their
positional encoding.}
\label{fig:more-co3d}
\end{figure}

\begin{figure}[t]
\centering
\includegraphics[width=\linewidth]{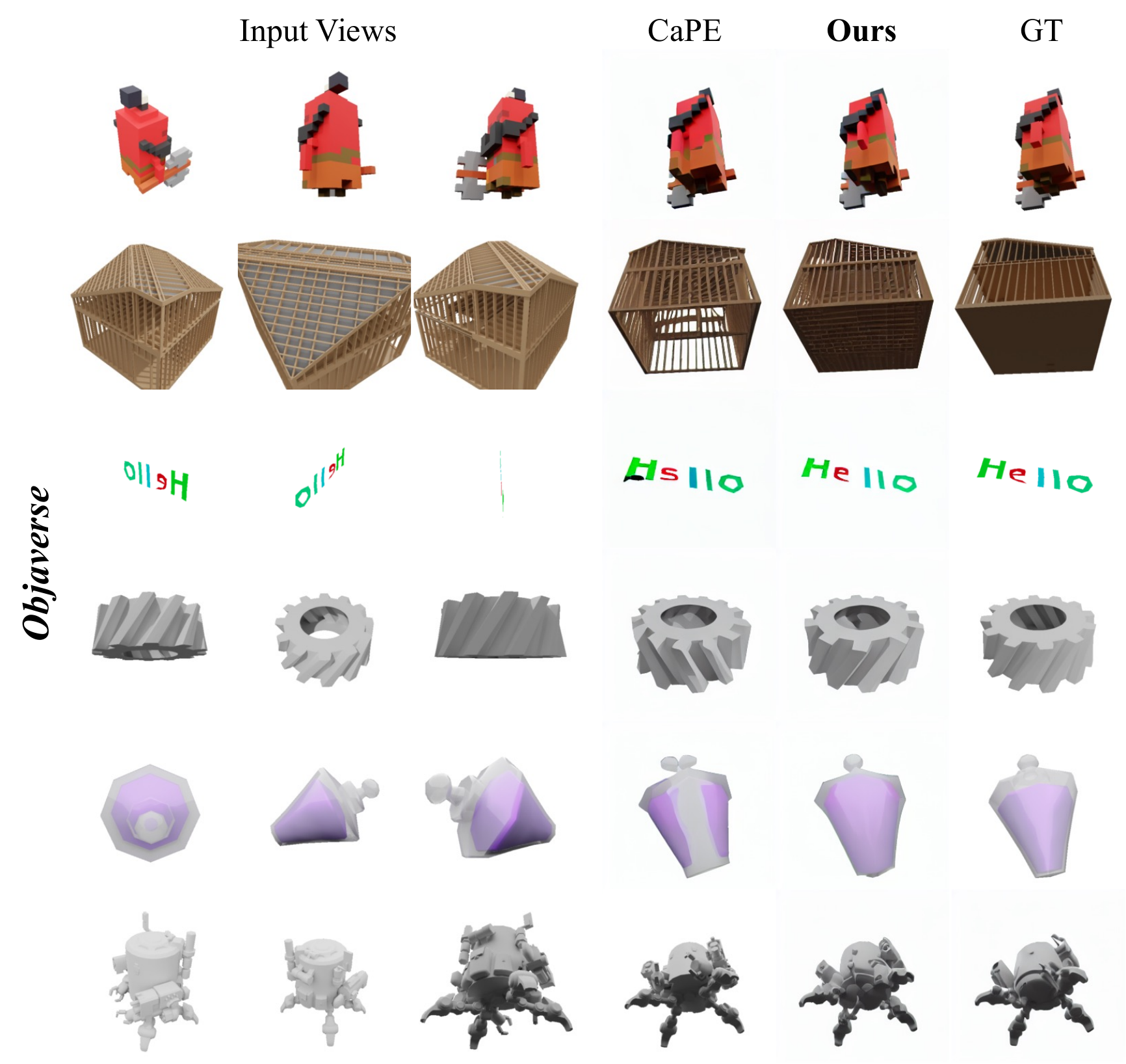}
\caption{\textbf{Additional qualitative comparison with EscherNet on Objaverse.}
Each row is one object. From left to right, the columns show three input views, the
novel views synthesized with CaPE and \ourpe{} (Ours), and the ground-truth target
view (GT). Both methods use the same EscherNet backbone and inputs and differ only
in their positional encoding.}
\label{fig:more-eschernet}
\end{figure}

Fig.~\ref{fig:more-re10k} and~\ref{fig:more-co3d} provide additional novel-view
comparisons against the baselines on RealEstate10K and CO3Dv2, complementing the
qualitative results in the main text. The comparisons
use the same LVSM decoder architecture and input views; the feature sources associated with each
encoding are specified in the main experiments. Consistent with the
quantitative results, \ourpe{} recovers sharper structure on the surfaces where a tilted
local plane matters most: oblique walls, floors, and table tops on RealEstate10K, and
the slanted faces and supporting surfaces of objects on CO3Dv2, where the projective
and ray-based encodings tend to blur the geometry or smear texture across depth
discontinuities. Fig.~\ref{fig:more-eschernet} further shows results under the
EscherNet setting on Objaverse, where \ourpe{} produces cleaner geometry and more
faithful appearance than CaPE from the same input views.

\end{document}